\documentclass{article}
\pdfoutput=1

\usepackage[numbers]{natbib}
\usepackage[preprint]{neurips_2018}
\usepackage[toc,page]{appendix}

\usepackage{color}

\usepackage{wrapfig}

\usepackage[utf8]{inputenc} 
\usepackage[T1]{fontenc}    
\usepackage{hyperref}       
\usepackage{url}            
\usepackage{booktabs}       
\usepackage{amsfonts}       
\usepackage{nicefrac}       
\usepackage{microtype}      

\usepackage{amsmath}

\DeclareMathOperator*{\argmin}{arg\,min}

\usepackage{graphicx,caption}
\usepackage[ruled,vlined,linesnumberedhidden]{algorithm2e}
\usepackage{mathtools}
\usepackage{makecell}
\usepackage{enumitem}

\DeclarePairedDelimiter{\ceil}{\lceil}{\rceil}
\DeclarePairedDelimiter{\floor}{\lfloor}{\rfloor}

\usepackage{bbm}
\newcommand{\x}{\mathbf{x}}

\newcommand{\D}{\mathcal{D}}

\newcommand{\E}{\mathbb{E}}
\newcommand{\R}{\mathbb{R}} 

\newcommand{\1}{\mathbbm{1}}
\newcommand{\Prob}{\mathrm{P}}

\newtheorem{prop}{Proposition}
\newtheorem{thm}{Theorem}
\newtheorem{lem}{Lemma}
\newenvironment{prf}{\noindent{\em Sketch of the proof\/}.}{{ $\Box$}\smallskip\par}
\newenvironment{myprf}{\noindent{\em Proof\/}.}{{ $\Box$}\smallskip\par}
\newenvironment{proof-prop}{\noindent{\em Proof of Theorem~\ref{prop:example}\/}.}{{ $\Box$}\smallskip\par}

\title{CatBoost: unbiased boosting with categorical features}

  \author{Liudmila Prokhorenkova$^{1,2}$, Gleb Gusev$^{1,2}$, Aleksandr  Vorobev$^{1}$, \\ \textbf{Anna Veronika Dorogush$^{1}$, Andrey  Gulin$^{1}$}\\
$^1$Yandex, Moscow, Russia \\
$^2$Moscow Institute of Physics and Technology, Dolgoprudny, Russia\\
\texttt{\{ostroumova-la, gleb57, alvor88, annaveronika, gulin\}@yandex-team.ru} \\
}

\begin{document}

\maketitle

\begin{abstract}
This paper presents the key algorithmic techniques behind CatBoost, a new gradient boosting toolkit.  Their combination leads to CatBoost outperforming other publicly available boosting implementations in terms of quality on a variety of datasets.   Two critical algorithmic advances introduced in CatBoost are the implementation of {\it ordered boosting}, a permutation-driven alternative to the classic algorithm, and an innovative algorithm for processing categorical features.  Both techniques were created to fight a {\it prediction shift} caused by a special kind of target leakage present in all currently existing implementations of gradient boosting algorithms. In this paper, we provide a detailed analysis of this problem and demonstrate that proposed algorithms solve it effectively, leading to excellent empirical results.
\end{abstract}

\section{Introduction}\label{sec:introduction}

Gradient boosting is a powerful machine-learning technique that achieves state-of-the-art results in a variety of practical tasks. For many years, it has remained the primary method for learning problems with heterogeneous features, noisy data, and complex dependencies: web search, recommendation systems, weather forecasting, and many others~\cite{caruana2006empirical,roe2005boosted,wu2010adapting,zhang2015gradient}. Gradient boosting is essentially a process of constructing an ensemble predictor by performing gradient descent in a functional space. It is backed by solid theoretical results that explain how strong predictors can be built by iteratively combining weaker models (\textit{base predictors}) in a greedy manner~\cite{kearns1994cryptographic}.

We show in this paper that all existing implementations of gradient boosting face the following statistical issue. A prediction model $F$ obtained after several steps of boosting relies on the targets of all training examples. We demonstrate that this actually leads to a shift of the distribution of $F(\x_k)\mid \x_k$ for a training example $\x_k$ from the distribution of $F(\x)\mid \x$ for a test example $\x$. This finally leads to a {\it prediction shift} of the learned model. We identify this problem as a special kind of target leakage in Section~\ref{sec:fighting}. Further, there is a similar issue in standard algorithms of preprocessing categorical features. One of the most effective ways~\cite{cestnik1990estimating,micci2001preprocessing} to use them in gradient boosting is converting categories to their target statistics. A target statistic is a simple statistical model itself, and it can also cause target leakage and a prediction shift. We analyze this in Section~\ref{sec:cat_features}.

In this paper, we propose \textit{ordering principle} to solve both problems. Relying on it, we derive \textit{ordered boosting}, a modification of standard gradient boosting algorithm, which avoids target leakage (Section~\ref{sec:fighting}), and a new algorithm for processing categorical features (Section~\ref{sec:cat_features}). Their combination is implemented as an open-source library\footnote{\url{https://github.com/catboost/catboost}} called CatBoost (for ``Categorical Boosting''), which outperforms the existing state-of-the-art implementations of gradient boosted decision trees~--- XGBoost~\cite{chen2016xgboost} and LightGBM~\cite{ke2017lightgbm}~--- on a diverse set of popular machine learning tasks (see Section~\ref{sec:experiments}). 

\section{Background}\label{sec:background}

Assume we observe a dataset of examples $\D=\{(\x_k,y_k)\}_{k=1..n}$, where $\x_k = (x_k^1, \ldots, x_k^m)$ is a random vector of $m$ {\it features} and $y_k\in \R$ is a {\it target}, which can be either binary or a numerical response. Examples $(\x_k,y_k)$ are independent and identically distributed according to some unknown distribution $P(\cdot, \cdot)$. The goal of a learning task is to train a function $F\colon \R^m\to \R$ which minimizes the expected loss $\mathcal{L}(F):=\E L(y, F(\x))$. Here $L(\cdot, \cdot)$ is a smooth loss function and $(\x,y)$ is a {\it test example} sampled from $P$ independently of the training set $\D$.

A gradient boosting procedure~\cite{friedman2001greedy} builds iteratively a sequence of {\it approximations} $F^t\colon \R^m\to \R$, $t=0,1,\ldots$ in a greedy fashion. Namely, $F^{t}$ is obtained from the previous approximation $F^{t-1}$ in an additive manner: $F^{t}=F^{t-1}+ \alpha h^{t}$, where $\alpha$ is a {\it step size} and function $h^t\colon \R^m\to \R$ (a {\it base predictor}) is chosen from a family of functions $H$ in order to minimize the expected loss:
\begin{equation}\label{boosting_step}
h^t = \argmin_{h\in H} \mathcal{L}(F^{t-1}+h)=\argmin_{h\in H}\E L(y, F^{t-1}(\x)+ h(\x)).
\end{equation}
The  minimization problem is usually approached by the {\it Newton method} using a second--order approximation of $\mathcal{L}(F^{t-1}+h^t)$ at $F^{t-1}$ or by taking a {\it (negative) gradient step}. Both methods are kinds of functional gradient descent~\cite{friedman2000additive,mason2000boosting}.  In particular, the gradient step $h^t$ is chosen in such a way that $h^t(\x)$ approximates $-g^t(\x,y)$, where $g^t(\x,y):=\frac{\partial L(y,s)}{\partial s}\big|_{s=F^{t-1}(\x)}$. Usually, the least-squares approximation is used:
\begin{equation}\label{eq:gradient_step}
h^t = \argmin_{h\in H} \E\left (-g^t(\x,y)- h(\x)\right )^2. 
\end{equation}
CatBoost is an implementation of gradient boosting, which uses binary decision trees as base predictors.  A {\it decision tree}~\cite{breiman1984classification,friedman2000additive,rokach2005top} is a model built by a recursive partition of the feature space $\R^m$ into several disjoint regions (tree nodes) according to the values of some {\it splitting attributes} $a$. Attributes are usually binary variables that identify that some feature $x^k$ exceeds some {\it threshold $t$}, that is, $a=\1_{\{x^k>t\}}$, where $x^k$ is either numerical or binary feature, in the latter case $t=0.5$.\footnote{Alternatively, non-binary splits can be used, e.g., a region can be split according to all values of a categorical feature. However, such splits, compared to binary ones, would lead to either shallow trees (unable to capture complex dependencies) or to very complex trees with exponential number of terminal nodes (having weaker target statistics in each of them). According to~\citep{breiman1984classification}, the tree complexity has a crucial effect on the accuracy of the model and less complex trees are less prone to overfitting.} Each final region (leaf of the tree) is assigned to a value, which is an estimate of the response $y$ in the region for the regression task or the predicted class label in the case of classification problem.\footnote{In a regression task, splitting attributes and leaf values are usually chosen by the least--squares criterion. Note that, in gradient boosting, a tree is constructed to approximate the negative gradient (see~Equation~\eqref{eq:gradient_step}), so it solves a regression problem.} In this way, a decision tree  $h$ can be written as
\begin{equation}\label{eq:leaf_values}
h(\x) = \sum_{j=1}^{J} b_j \1_{\{\x \in R_j\}},
\end{equation}
where $R_j$ are the disjoint regions corresponding to the leaves of the tree.  

\section{Categorical features}\label{sec:cat_features}

\subsection{Related work on categorical features}\label{CatFeatures}

A categorical feature is one with a discrete set of values called \textit{categories} that are not comparable to each other. One popular technique for dealing with categorical features in boosted trees is \textit{one-hot encoding} \cite{ClickFeatures_1,micci2001preprocessing}, i.e., for each category, adding a new binary feature indicating it. However, in the case of high cardinality features (like, e.g., ``user ID'' feature), such technique leads to infeasibly large number of new features. To address this issue, one can group categories into a limited number of clusters and then apply one-hot encoding. A popular method is to group categories by \textit{target statistics} (TS) that estimate expected target value in each category. Micci-Barreca~\cite{micci2001preprocessing} proposed to consider TS as a new numerical feature instead. Importantly, among all possible partitions of categories into two sets, an optimal split on the training data in terms of logloss, Gini index, MSE can be found among thresholds for the numerical TS feature~\cite[Section 4.2.2]{breiman1984classification}~\cite[Section 9.2.4]{GradientBasedStat}. In LightGBM~\cite{LightGBM_website}, categorical features are converted to gradient statistics at each step of gradient boosting. Though providing important information for building a tree, this approach can dramatically increase (i)~computation time, since it calculates statistics for each categorical value at each step, and (ii)~memory consumption to store which category belongs to which node for each split based on a categorical feature. To overcome this issue, LightGBM groups tail categories into one cluster~\cite{LightGBM_code} and thus looses part of information. Besides, the authors claim that it is still better to convert categorical features with high cardinality to numerical features~\cite{LightGBM_website_2}. Note that TS features require calculating and storing only one number per one category.

Thus, using TS as new numerical features seems to be the most efficient method of handling categorical features with minimum information loss. TS are widely-used, e.g., in the click prediction task (click-through rates)~\cite{OnlineLearning_1,OnlineLearning_4,OnlineLearning_2,OnlineLearning_3}, where such categorical features as user, region, ad, publisher play a crucial role. We further focus on ways to calculate TS and leave one-hot encoding and gradient statistics out of the scope of the current paper. At the same time, we believe that the ordering principle proposed in this paper is also effective for gradient statistics. 

\subsection{Target statistics}\label{sec:TS}

As discussed in Section~\ref{CatFeatures}, an effective and efficient way to deal with a categorical feature $i$ is to substitute the category $x_{k}^i$ of $k$-th training example with \textit{one} numeric feature equal to some \textit{target statistic} (TS) $\hat x_k^i$. Commonly, it estimates the expected target $y$ conditioned by the category: $\hat x_k^i\approx \E (y \mid x^i = x_k^i)$.  

\paragraph{Greedy TS}
A straightforward approach is to estimate $\E (y \mid x^i = x_k^i)$ as the average value of $y$ over the training examples with the same category $x_k^i$~\cite{micci2001preprocessing}. This estimate is noisy for low-frequency categories, and one usually smoothes it by some prior $p$:
\begin{equation}\label{eq:ctr_greedy_prior}
\hat x_{k}^{i} = \frac{\sum_{j=1}^n{\1_{\{x_{j}^{i}=x_{k}^{i}\}}\cdot y_j}+a\,p}{\sum_{j=1}^n\1_{\{x_{j}^{i}=x_{k}^{i}\}}+a}\,,
\end{equation}
where $a>0$ is a parameter. A common setting for $p$ is the average target value in the dataset~\cite{micci2001preprocessing}. 

The problem of such \textit{greedy} approach is \textit{target leakage}:  feature $\hat x_k^i$ is computed using $y_k$, the target of $\x_k$. This leads to a conditional shift~\cite{zhang2013domain}: the distribution of~$\hat x^{i}|y$  differs for training and test examples. The following extreme example illustrates how dramatically this may affect the generalization error of the learned model.  Assume $i$-th feature is categorical, all its values are unique, and for each category $A$, we have $\Prob(y=1\mid x^i=A) = 0.5$ for a classification task. Then, in the training dataset, $\hat x_{k}^{i} = \frac{y_k+ap}{1+a}$, so it is sufficient to make only one split with threshold $t=\frac{0.5+ap}{1+a}$  to perfectly classify all training examples. However, for all test examples, the value of the greedy TS is $p$, and the obtained model predicts $0$ for all of them if $p<t$ and predicts $1$ otherwise, thus having accuracy $0.5$ in both cases. To this end, we formulate the following desired property for TS: 
\begin{itemize}[noitemsep,nolistsep]
\item[P1] \textit{$\E(\hat x^i\mid y=v)$ = $\E(\hat x_k^i \mid y_k=v)$, where $(\x_k,y_k)$ is the $k$-th training example}. 
\end{itemize}
In our example above, $\E(\hat x_k^i \mid y_k) = \frac{y_k+ap}{1+a}$ and $\E(\hat x^i\mid y) = p$ are different. 

There are several ways to avoid this conditional shift. Their general idea is to compute the TS for $\x_k$ on a subset of examples $\D_k \subset \D\setminus\{\x_k\}$  excluding $\x_k$:
\begin{equation}\label{eq:general_ctr}
\hat x_{k}^{i} = \frac{\sum_{\x_j \in \D_k}{\1_{\{x_{j}^{i}=x_{k}^{i}\}}\cdot y_j}+a\,p}{\sum_{\x_j \in \D_k}\1_{\{x_{j}^{i}=x_{k}^{i}\}}+a}\,.
\end{equation}

\paragraph{Holdout TS}  One way is to partition the training dataset into two parts $\D = \hat \D_0 \sqcup \hat \D_1$ and use $\D_k = \hat \D_0$ for calculating the TS according to~\eqref{eq:general_ctr} and $\hat \D_1$ for training (e.g., applied in~\cite{chen2016xgboost} for Criteo dataset). Though such \textit{holdout} TS satisfies P1, this approach significantly reduces the amount of data used both for training the model and calculating the TS. So, it violates the following desired property:
\begin{itemize}[noitemsep,nolistsep]
\item[P2] \textit{Effective usage of all training data for calculating TS features and for learning a model.}
\end{itemize}

\paragraph{Leave-one-out TS} At first glance, a \textit{leave-one-out} technique might work well: take $\D_k = \D \setminus \x_k$ for training examples $\x_k$ and $\D_k = \D$ for test ones~\cite{TopKaggler}. Surprisingly, it does not prevent target leakage. Indeed, consider a constant categorical feature: $x_k^i = A$ for all examples. Let $n^+$ be the number of examples with $y=1$, then $\hat x_{k}^{i} = \frac{n^+ - y_k + a\,p}{n - 1 + a}$ and one can perfectly classify the training dataset by making a split with threshold $t = \frac{n^+ - 0.5 + a\,p}{n - 1 + a}$.

\paragraph{Ordered TS}
CatBoost uses a more effective strategy. It relies on the ordering principle, the central idea of the paper, and is inspired by online learning algorithms which get training examples sequentially in time~\cite{OnlineLearning_1,OnlineLearning_4,OnlineLearning_2,OnlineLearning_3}). Clearly, the values of TS for each example rely only on the observed history. To adapt this idea to standard offline setting, we introduce an artificial ``time'', i.e., a random permutation $\sigma$ of the training examples. Then, for each example, we use all the available ``history'' to compute its TS, i.e., take $\D_k = \{\x_j:\sigma(j)<\sigma(k)$\} in Equation~\eqref{eq:general_ctr} for a training example and $\D_k = \D$ for a test one. The obtained \textit{ordered} TS satisfies the requirement P1 and allows to use all training data for learning the model (P2). Note that, if we use only one random permutation, then preceding examples have TS with much higher variance than subsequent ones. To this end, CatBoost uses different permutations for different steps of gradient boosting, see details in Section~\ref{sec:implementation}.

\section{Prediction shift and ordered boosting}\label{sec:fighting}

\subsection{Prediction shift}
\label{sec:biasness_definition}

In this section, we reveal the problem of prediction shift in gradient boosting, which was neither recognized nor previously addressed. Like in case of TS, prediction shift is caused by a special kind of target leakage. Our solution is called \textit{ordered boosting} and resembles the ordered TS method.

Let us go back to the gradient boosting procedure described in Section~\ref{sec:background}. In practice, the expectation in~\eqref{eq:gradient_step} is unknown  and is usually approximated using the same dataset $\D$: 
\begin{equation}\label{eq:gradient_approximation}
h^t = \argmin_{h\in H} \frac{1}{n}\sum_{k=1}^{n} \left (-g^t(\x_k, y_k) -  h(\x_k)\right )^2.
\end{equation}
Now we describe and analyze the following chain of shifts:
\begin{enumerate}[noitemsep,nolistsep]
\item the conditional distribution of the gradient $g^t(\x_k, y_k)\mid \x_k$ (accounting for randomness of $\D\setminus\{\x_k\}$) is shifted from that distribution on a test example $g^t(\x, y)\mid \x$;
\item in turn, base predictor $h^t$ defined by Equation~\eqref{eq:gradient_approximation} is biased from the solution of Equation~\eqref{eq:gradient_step};
\item this, finally, affects the generalization ability of the trained model $F^t$.
\end{enumerate}
As in the case of TS, these problems are caused by the target leakage.  Indeed, gradients used at each step are estimated using the target values of the same data points the current model $F^{t-1}$ was built on. However, the conditional distribution $F^{t-1}(\x_k)\mid \x_k$ for a training example $\x_k$ is shifted, in general, from the distribution $F^{t-1}(\x)\mid \x$ for a test example $\x$. We call this a {\it prediction shift}.

\paragraph{Related work on prediction shift} 
The shift of gradient conditional distribution $g^t(\x_k, y_k)\mid \x_k$ was previously mentioned in papers on boosting~\cite{breiman2001using,friedman2002stochastic} but was not formally defined. Moreover, even the existence of non-zero shift was not proved theoretically. Based on the out-of-bag estimation~\cite{breiman1996out}, Breiman proposed \textit{iterated bagging}~\cite{breiman2001using} which constructs a bagged weak learner at each iteration on the basis of ``out-of-bag'' residual estimates. However, as we formally show in Appendix~\ref{appendix::iterated_bagging}, such residual estimates are still shifted. Besides, the bagging scheme increases learning time by factor of the number of data buckets. Subsampling of the dataset at each iteration proposed by Friedman~\cite{friedman2002stochastic} addresses the problem much more heuristically and also only alleviates it.

\paragraph{Analysis of prediction shift} 
We formally analyze the problem of prediction shift in a simple case of a regression task with the quadratic loss function $L(y,\hat{y})=(y-\hat{y})^2$.\footnote{We restrict the rest of Section~\ref{sec:fighting} to this case, but the approaches of Section~\ref{sec:ordered_boosting} are applicable to other tasks.} In this case, the negative gradient $-g^t(\x_k, y_k)$ in Equation~\eqref{eq:gradient_approximation} can be substituted by the residual function $r^{t-1}(\x_k,y_k):= y_k-F^{t-1}(\x_k)$.\footnote{Here we removed the multiplier 2, what does not matter for further analysis.} Assume we have $m=2$ features $x^1, x^2$ that are i.i.d. Bernoulli random variables with $p = 1/2$ and $y=f^*(\x)=c_1x^1+c_2x^2$. Assume we make $N=2$ steps of gradient boosting with decision stumps (trees of depth 1) and step size $\alpha = 1$. We obtain a model $F=F^2=h^1+h^2$. W.l.o.g., we assume that $h^1$ is based on $x^1$ and $h^2$ is based on $x^2$, what is typical for $|c_1|>|c_2|$ (here we set some asymmetry between $x^1$ and $x^2$). 
\begin{thm}\label{prop:example}
1. If two independent samples $\D_1$ and $\D_2$ of size $n$ are used to estimate $h^1$ and $h^2$, respectively, using Equation~\eqref{eq:gradient_approximation}, then $\E_{\D_1, \D_2} F^2(\x)=f^*(\x)+O(1/2^{n})$ for any $\x\in \{0,1\}^2$. \\
2. If the same dataset $\D = \D_1 = \D_2$ is used in Equation~\eqref{eq:gradient_approximation} for both $h^1$ and $h^2$, then $\E_{\D} F^2(\x)=f^*(\x)-\frac{1}{n-1}c_2(x^2-\frac{1}{2})+O(1/2^{n})$.
\end{thm}

This theorem means that the trained model is an unbiased estimate of the true dependence $y=f^*(\x)$, when we use independent datasets at each gradient step.\footnote{Up to an exponentially small term, which occurs for a technical reason.} On the other hand, if we use the same dataset at each step, we suffer from a bias $-\frac{1}{n-1}c_2(x^2-\frac{1}{2})$, which is inversely proportional to the data size $n$. Also, the value of the bias can depend on the relation $f^*$: in our example, it is proportional to $c_2$. We track the chain of shifts for the second part of Theorem~\ref{prop:example} in a sketch of the proof below,  while the full proof of Theorem~\ref{prop:example} is available in Appendix~\ref{appendix::proof}.

\begin{prf}
Denote by $\xi_{st}$, $s,t\in \{0,1\}$, the number of examples $(\x_k,y_k)\in\D$ with $\x_k=(s,t)$. We have $h^1(s,t)=c_1s + \frac{c_2\xi_{s 1}}{\xi_{s 0}+\xi_{s 1}}$. Its expectation $\E(h^1(\x))$ on a test example $\x$ equals $c_1x^1+\frac{c_2}{2}$. At the same time, the expectation $\E(h^1(\x_k))$ on a training example $\x_k$ is different and equals $(c_1x^1+\frac{c_2}{2}) - c_2(\frac{2x^2-1}{n}) +O(2^{-n})$. That is, we experience a prediction shift of $h^1$. As a consequence, the expected value of $h^2(\x)$ is $\E(h^2(\x))=c_2(x^2-\frac{1}{2}) (1-\frac{1}{n-1}) +O(2^{-n})$ on a test example $\x$ and $\E(h^1(\x)+h^2(\x))=f^*(\x)-\frac{1}{n-1}c_2(x^2-\frac{1}{2})+O(1/2^{n})$. 
\end{prf}

Finally, recall that greedy TS $\hat x^i$ can be considered as a simple statistical model predicting the target $y$ and it suffers from a similar problem, conditional shift of $\hat x^i_k\mid y_k$, caused by the target leakage, i.e., using $y_k$ to compute $\hat x^i_k$.

\subsection{Ordered boosting}\label{sec:ordered_boosting}

Here we propose a boosting algorithm which does not suffer from the prediction shift problem described in Section~\ref{sec:biasness_definition}. Assuming access to an unlimited amount of training data, we can easily construct such an algorithm. At each step of boosting, we sample a new dataset $\D_t$ independently and obtain unshifted residuals by applying the current model to new training examples. In practice, however, labeled data is limited. Assume that we learn a model with $I$ trees. To make the residual $r^{I-1}(\x_k,y_k)$ unshifted, we need to have $F^{I-1}$ trained without the example $\x_k$. Since we need unbiased residuals for all training examples, no examples may be used for training $F^{I-1}$, which at first glance makes the training process impossible. However, it is possible to maintain a set of models differing by examples used for their training. Then, for calculating the residual on an example, we use a model trained without it. In order to construct such a set of models, we can use the ordering principle previously applied to TS in Section~\ref{sec:TS}. To illustrate the idea, assume  that we take one random permutation $\sigma$ of the training examples and maintain $n$ different {\it supporting}  models $M_1, \ldots, M_n$ such that the model $M_i$ is learned using only the first $i$ examples in the permutation. At each step, in order to obtain the residual for $j$-th sample, we use the model $M_{j-1}$ (see Figure~\ref{fig:comics}). The resulting Algorithm~\ref{alg:ordered} is called \textit{ordered boosting} below. Unfortunately, this algorithm is not feasible in most practical tasks due to the need of training $n$ different models,  what increase the complexity and memory requirements by $n$ times. In CatBoost, we implemented a modification of this algorithm on the basis of the gradient boosting algorithm with decision trees as base predictors (GBDT) described in Section~\ref{sec:implementation}.

\paragraph{Ordered boosting with categorical features} 

In Sections~\ref{sec:TS} and~\ref{sec:ordered_boosting} we proposed to use random permutations $\sigma_{cat}$ and $\sigma_{boost}$ of training examples for the TS calculation and for ordered boosting, respectively. Combining them in one algorithm, we should take $\sigma_{cat} = \sigma_{boost}$ to avoid prediction shift. This guarantees that target $y_i$ is not used for training $M_{i}$ (neither for the TS calculation, nor for the gradient estimation). See Appendix~\ref{appendix::combination} for theoretical guarantees. Empirical results confirming the importance of having $\sigma_{cat} = \sigma_{boost}$ are presented in Appendix~\ref{appendix::experimental_res}.

\section{Practical implementation of ordered boosting}\label{sec:implementation}

CatBoost has two boosting modes, \textit{Ordered} and \textit{Plain}. The latter mode is the standard GBDT algorithm with inbuilt ordered TS. The former mode presents an efficient modification of Algorithm~\ref{alg:ordered}. A formal description of the algorithm is included in Appendix~\ref{appendix::full_description}. In this section, we overview the most important implementation details.

\begin{minipage}{0.4\textwidth}
\centering
\includegraphics[width=\textwidth]{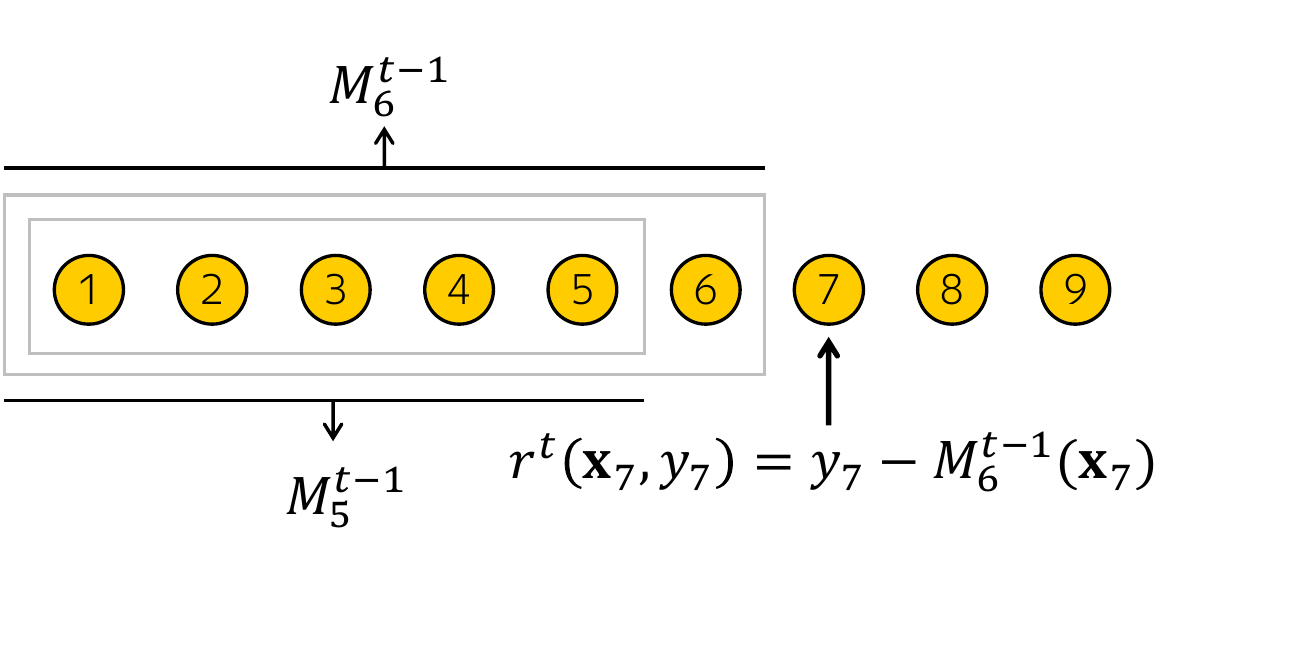}
\captionof{figure}{Ordered boosting principle,\\ examples are ordered according to $\sigma$.}\label{fig:comics}
\vspace{1.3cm}
\begin{algorithm}[H]
\SetKwInOut{Input}{input}\SetKwInOut{Output}{output}
  \Input{\,\,$\{(\x_k,y_k)\}_{k=1}^n$, $I$\;}
  \BlankLine
  $\sigma \leftarrow$ random permutation of $[1,n]$ \;
  $M_i \leftarrow 0$ for $i = 1..n$\;
  \For{$t \leftarrow 1$ \KwTo $I$}{ 
      \For{$i \leftarrow 1$ \KwTo $n$}{
          $r_i \leftarrow y_i - M_{\sigma(i)-1}(\x_i)$\;
      }
      \For{$i \leftarrow 1$ \KwTo $n$}{
          $\Delta M \leftarrow LearnModel((\x_j,r_j): \sigma(j) \le i)$\;
          $M_i \leftarrow M_i + \Delta M$ \;
      } 
      }
       \Return{$M_n$}
  {\caption{Ordered boosting}
  \label{alg:ordered}}
\end{algorithm}
\end{minipage}
\begin{minipage}{0.58\textwidth}
\begin{algorithm}[H]
\SetKwInOut{Input}{input}\SetKwInOut{Output}{output}
  \Input{\,\,$M$, $\{(\x_i,y_i)\}_{i=1}^n$, $\alpha$, $L$, $\{\sigma_i\}_{i=1}^s$, $Mode$ }
  \BlankLine 
  $grad \leftarrow CalcGradient(L,M,y)$\;
  $r \leftarrow random(1,s)$\;
  \If{$Mode = Plain$}{
     $G \leftarrow (grad_{r}(i) \mbox{ for } i=1..n)$\;}
  \If{$Mode=Ordered$}{
    $G \leftarrow (grad_{r, \sigma_r(i)-1}(i) \mbox{ for } i=1..n)$\;}
  $T \leftarrow$ empty tree\;
  \ForEach{step of top-down procedure}{
   \ForEach{candidate split $c$ }{
      $T_c \leftarrow$ add split $c$ to $T$\;
      \If{$Mode = Plain$}{
              $\Delta(i) \leftarrow \mathrm{avg}(grad_{r}(p)$ for $p:\ leaf_r(p)=leaf_r(i) )$ \ for $i=1..n$\; }
      \If{$Mode=Ordered$}{
              $\Delta(i) \leftarrow \mathrm{avg}(grad_{r, \sigma_r(i)-1}(p)$ for $p:\  leaf_r(p)=leaf_r(i), \sigma_r(p)<\sigma_r(i))$ \ for $i=1..n$\;}
       $loss(T_c) \leftarrow \cos(\Delta,G)$
  }
   $T \leftarrow \argmin_{T_c}(loss(T_c))$
   }
  \If{$Mode = Plain$}{
     $M_{r'}(i) \leftarrow M_{r'}(i) - \alpha \, \mathrm{avg}(grad_{r'}(p)$ for $p:\ leaf_{r'}(p)=leaf_{r'}(i) )$ for $r'=1..s$, $i=1..n$\; }
  \If{$Mode=Ordered$}{
    $M_{r',j}(i) \leftarrow M_{r',j}(i) - \alpha \, \mathrm{avg}(grad_{r',j}(p)$ for $p:\  leaf_{r'}(p)=leaf_{r'}(i), \sigma_{r'}(p)\leq j)$ for $r'=1..s$, $i=1..n$, $j\geq \sigma_{r'}(i)-1$\;
} 
  \Return{$T, M$}
  {\caption{Building a tree in CatBoost}
      \label{alg:CatBoostStepMain}}
      \end{algorithm} 
\end{minipage}
\hspace{0.5cm}

At the start, CatBoost generates $s+1$ independent random permutations of the training dataset. The permutations $\sigma_1, \ldots, \sigma_s$ are used for evaluation of splits that define tree structures (i.e., the internal nodes), while $\sigma_0$ serves for choosing the leaf values $b_j$ of the obtained trees (see Equation~\eqref{eq:leaf_values}). For examples with short history in a given permutation, both TS and predictions used by ordered boosting ($M_{\sigma(i)-1}(\x_i)$ in Algorithm~\ref{alg:ordered}) have a high variance. Therefore, using only one permutation may increase the variance of the final model predictions, while several permutations allow us to reduce this effect in a way we further describe. The advantage of several permutations is confirmed by our experiments in Section~\ref{sec:n_permut}.

\paragraph{Building a tree} 
In CatBoost, base predictors are oblivious decision trees~\cite{ferov2016enhancing, gulin2011winning} also called decision tables~\cite{BoostedDecisionTables}. Term oblivious means that the same splitting criterion is used across an entire level of the tree. Such trees are balanced, less prone to overfitting, and allow speeding up execution at testing time significantly. The procedure of building a tree in CatBoost is described in Algorithm~\ref{alg:CatBoostStepMain}.

In the Ordered boosting mode, during the learning process, we maintain the supporting models $M_{r,j}$, where $M_{r,j}(i)$ is the current prediction for the $i$-th example based on the first $j$ examples in the permutation~$\sigma_r$. At each iteration~$t$ of the algorithm, we sample a random permutation $\sigma_r$ from $\{\sigma_1, \ldots, \sigma_s\}$ and construct a tree~$T_t$ on the basis of it. First, for categorical features, all TS are computed according to this permutation. Second, the permutation affects the tree learning procedure. Namely, based on $M_{r,j}(i)$, we compute the corresponding gradients $grad_{r,j}(i)=\frac{\partial L(y_i,s)}{\partial s}\big|_{s=M_{r,j}(i)}$. Then, while constructing a tree, we approximate the gradient $G$ in terms of the cosine similarity 
$\cos(\cdot,\cdot)$, where, for each example $i$, we take the gradient $grad_{r,\sigma(i)-1}(i)$ (it is based only on the previous examples in $\sigma_r$). At the candidate splits evaluation step, the leaf value~$\Delta(i)$ for example~$i$ is obtained individually by averaging the gradients $grad_{r,\sigma_r(i)-1}$ of the preceding examples $p$ lying in the same leaf $leaf_r(i)$ the example $i$ belongs to. Note that $leaf_r(i)$ depends on the chosen permutation $\sigma_r$, because $\sigma_r$ can influence the values of ordered TS for example~$i$. When the tree structure $T_t$ (i.e., the sequence of splitting attributes) is built, we use it to boost all the models $M_{r',j}$. Let us stress that {\it one common} tree structure~$T_t$ is used for all the models, but this tree is added to different $M_{r',j}$ with different sets of leaf values depending on $r'$ and $j$, as described in Algorithm~\ref{alg:CatBoostStepMain}.

The Plain boosting mode works similarly to a standard GBDT procedure, but, if categorical features are present, it maintains $s$ supporting models $M_{r}$ corresponding to TS based on $\sigma_1, \ldots, \sigma_s$. 

\paragraph{Choosing leaf values} 
Given all the trees constructed, the leaf values of the final model $F$ are calculated by the standard gradient boosting procedure equally for both modes. Training examples~$i$ are matched to leaves $leaf_0(i)$, i.e., we use permutation $\sigma_0$ to calculate TS here. When the final model $F$ is applied to a new example at testing time, we use TS calculated on the whole training data according to Section~\ref{sec:TS}.

\paragraph{Complexity}
In our practical implementation, we use one important trick, which significantly reduces the computational complexity of the algorithm. Namely, in the Ordered mode, instead of $O(s\, n^2)$ values $M_{r,j}(i)$, we store and update only the values $M'_{r,j}(i): = M_{r,2^j}(i)$ for $j=1,\ldots,\ceil{\log_2 n}$ and all $i$ with $\sigma_r(i) \le 2^{j+1}$, what reduces the number of maintained supporting predictions to $O(s\,n)$. 
See Appendix~\ref{appendix::full_description} for the pseudocode of this modification of Algorithm~\ref{alg:CatBoostStepMain}.

In Table~\ref{tab:complexity}, we present the computational complexity of different components of both CatBoost modes per one iteration (see Appendix~\ref{appendix::time_complexity_theory} for the proof). Here $N_{TS,t}$ is the number of TS to be calculated at the iteration $t$ and $C$ is the set of candidate splits to be considered at the given iteration. It follows that our implementation of ordered boosting with decision trees has the same asymptotic complexity as the standard GBDT with ordered TS. In comparison with other types of TS (Section~\ref{sec:TS}), ordered TS slow down by $s$ times the procedures $CalcGradient$, updating supporting models $M$, and computation of~TS.

\begin{table}[t]
\caption{Computational complexity.}
\label{tab:complexity}
\centering
\small
\begin{tabular}{lccccc}
  \toprule
 Procedure & CalcGradient & Build $T$ & Calc all $b^t_j$ & Update $M$ & Calc ordered TS  \\
  \midrule
 Complexity for iteration $t$ 
 &  $O(s\cdot n)$ &  $O(|C|\cdot n)$ & $O(n)$  &   $O(s\cdot n)$ & $O(N_{TS,t}\cdot n)$\\
  \bottomrule
\end{tabular}
\end{table}

\paragraph{Feature combinations}
Another important detail of CatBoost is using combinations of categorical features as additional categorical features which capture high-order dependencies like joint information of user ID and ad topic in the task of ad click prediction. The number of possible combinations grows exponentially with the number of categorical features in the dataset, and it is infeasible to process all of them. CatBoost constructs combinations in a greedy way. Namely, for each split of a tree, CatBoost combines (concatenates) all categorical features (and their combinations) already used for previous splits in the current tree with all categorical features in the dataset. Combinations are converted to TS on the fly. 

\paragraph{Other important details}
Finally, let us discuss two options of the CatBoost algorithm not covered above. The first one is subsampling of the dataset at each iteration of boosting procedure, as proposed by Friedman~\cite{friedman2002stochastic}. We claimed earlier in Section~\ref{sec:biasness_definition} that this approach alone cannot fully avoid the problem of prediction shift. However, since it has proved effective, we implemented it in both modes of CatBoost as a Bayesian bootstrap procedure. Specifically, before training a tree according to Algorithm~\ref{alg:CatBoostStepMain}, we assign a weight $w_i=a_i^t$ to each example $i$, where $a_i^t$ are generated according to the Bayesian bootstrap procedure (see~\cite[Section 2]{rubin1981bayesian}). These weights are used as multipliers for gradients $grad_r(i)$ and $grad_{r,j}(i)$, when we calculate $\Delta(i)$ and the components of the vector $\Delta-G$ to define $loss(T_c)$. 

The second option deals with first several examples in a permutation. For examples $i$ with small values $\sigma_{r}(i)$, the variance of $grad_{r,\sigma_r(i)-1}(i)$ can be high. Therefore, we discard $\Delta(i)$ from the beginning of the permutation, when we calculate $loss(T_c)$ in Algorithm~\ref{alg:CatBoostStepMain}. Particularly, we eliminate the corresponding components of vectors $G$ and $\Delta$ when calculating the cosine similarity between them. 

\section{Experiments}\label{sec:experiments}

\paragraph{Comparison with baselines}

We compare our algorithm with the most popular open-source libraries~--- XGBoost and LightGBM~--- on several well-known machine learning tasks. The detailed description of the experimental setup together with dataset descriptions is available in Appendix~\ref{appendix::experimental_setup}. The source code of the experiment is available, and the results can be reproduced.\footnote{\url{https://github.com/catboost/benchmarks/tree/master/quality_benchmarks}} For all learning algorithms, we preprocess categorical features using the ordered TS method described in Section~\ref{sec:TS}. The parameter tuning and training were performed on 4/5 of the data and the testing was performed on the remaining 1/5.\footnote{For Epsilon, we use default parameters instead of parameter tuning due to large running time for all algorithms. We tune only the number of trees to avoid overfitting.} The results measured by logloss and zero-one loss are  presented in Table~\ref{tab:baselines} (the absolute values for the baselines are in Appendix~\ref{appendix::experimental_res}). For CatBoost, we used Ordered boosting mode in this experiment.\footnote{The numbers for CatBoost in Table~\ref{tab:baselines} may slightly differ from the corresponding numbers in our GitHub repository, since we use another version of CatBoost with all the discussed features implemented.} One can see that CatBoost outperforms other algorithms on all the considered datasets. We also measured statistical significance of improvements presented in Table~\ref{tab:baselines}: except three datasets (Appetency, Churn and Upselling) the improvements are statistically significant with p-value $\ll 0.01$ measured by the paired one-tailed t-test. 

To demonstrate that our implementation of plain boosting is an appropriate baseline for our research, we show that a \textit{raw setting} of CatBoost provides state-of-the-art quality. Particularly, we take a setting of CatBoost, which is close to classical GBDT~\cite{friedman2001greedy}, and compare it with the baseline boosting implementations in Appendix~\ref{appendix::experimental_res}. Experiments show that this raw setting differs from the baselines insignificantly.

\begin{minipage}{0.49\textwidth}
\vspace{5pt}
\captionof{table}{Comparison with baselines: logloss / zero-one loss (relative increase for baselines).}\label{tab:baselines}
\vspace{6pt}
\scriptsize
\centering
\begin{tabular}{lccc}
  \cmidrule{2-4}
 & CatBoost &  LightGBM &  XGBoost \\
   \midrule
Adult & 
\textbf{0.270 / 0.127}  & 
+2.4\% / +1.9\% &  
+2.2\% / +1.0\%  \\
Amazon & 
\textbf{0.139 / 0.044} & 
+17\% / +21\% & 
+17\% / +21\% \\
Click & 
\textbf{0.392 / 0.156} & 
+1.2\% / +1.2\%  & 
+1.2\% / +1.2\% \\
Epsilon & 
\textbf{0.265 / 0.109} & 
+1.5\% / +4.1\% & 
+11\% / +12\% \\
Appetency & 
\textbf{0.072 / 0.018} & 
+0.4\% / +0.2\% & 
+0.4\%  / +0.7\% \\
Churn & 
\textbf{0.232 / 0.072}  & 
+0.1\% / +0.6\% &  
+0.5\% / +1.6\%  \\
Internet & 
\textbf{0.209 / 0.094} & 
+6.8\% / +8.6\% &
+7.9\% / +8.0\% \\
Upselling & 
\textbf{0.166 / 0.049}  & 
+0.3\% / +0.1\%	& 
+0.04\% / +0.3\% \\
Kick & 
\textbf{0.286 / 0.095}  & 
+3.5\% / +4.4\% &
+3.2\% / +4.1\%  \\
\bottomrule
\end{tabular}
\end{minipage}
\ \ \ \ \  \ 
\begin{minipage}{0.45\textwidth}\captionof{table}{Plain boosting mode: logloss, zero-one loss and their change relative to Ordered boosting mode.}\label{tab:plain}
\centering
\scriptsize
\begin{tabular}{lcc}
  \cmidrule{2-3}
 & Logloss &  Zero-one loss \\
   \midrule
Adult & 
0.272 (+1.1\%) &
0.127	(-0.1\%) \\
Amazon & 
0.139 (-0.6\%) & 
0.044	(-1.5\%) \\	
Click & 
0.392	(-0.05\%) & 
0.156	(+0.19\%) \\
Epsilon & 
0.266 (+0.6\%) & 
0.110 (+0.9\%) \\
Appetency & 
0.072	(+0.5\%) & 
0.018	(+1.5\%) \\
Churn & 
0.232	(-0.06\%) & 
0.072	(-0.17\%)  \\
Internet & 
0.217 (+3.9\%) & 
0.099 (+5.4\%) \\
Upselling & 
0.166 (+0.1\%) & 
0.049 (+0.4\%) \\
Kick & 
0.285 (-0.2\%) & 
0.095 (-0.1\%) \\
\bottomrule
\end{tabular}
\end{minipage}

We also empirically analyzed the running times of the algorithms on Epsilon dataset. The details of the comparison can be found in Appendix~\ref{appendix::time_complexity_experiments}. To summarize, we obtained that CatBoost Plain and LightGBM are the fastest ones followed by Ordered mode, which is about 1.7 times slower. 

\paragraph{Ordered and Plain modes}

In this section, we compare two essential boosting modes of CatBoost: Plain and Ordered. First, we compared their performance on all the considered datasets, the results are presented in Table~\ref{tab:plain}. It can be clearly seen that Ordered mode is particularly useful on small datasets. Indeed, the largest benefit from Ordered is observed on Adult and Internet datasets, which are relatively small (less than 40K training examples), which supports our hypothesis that a higher bias negatively affects the performance. Indeed, according to Theorem~\ref{prop:example} and our reasoning in Section~\ref{sec:biasness_definition}, bias is expected to be larger for smaller datasets (however, it can also depend on other properties of the dataset, e.g., on the dependency between features and target). In order to further validate this hypothesis, we make the following experiment: we train CatBoost in Ordered and Plain modes on randomly filtered datasets and compare the obtained losses, see Figure~\ref{fig:plain_filtered}. As we expected, for smaller datasets the relative performance of Plain mode becomes worse. To save space, here we present the results only for logloss; the figure for zero-one loss is similar.

We also compare Ordered and Plain modes in the above-mentioned raw setting of CatBoost in Appendix~\ref{appendix::experimental_res} and conclude that the advantage of Ordered mode is not caused by interaction with specific CatBoost options.

\begin{minipage}{0.52\textwidth}\captionof{table}{Comparison of target statistics, relative change in logloss / zero-one loss compared to ordered TS.}\label{tab:ctr}
\centering
\scriptsize
\begin{tabular}{lccc}
  \cmidrule{2-4}
 & Greedy & Holdout & Leave-one-out 
 \\
   \midrule
Adult & 
+1.1\% / +0.8\% &
+2.1\% / +2.0\% &
+5.5\% / +3.7\% 
\\
Amazon & 
+40\% / +32\% &
+8.3\% / +8.3\% &
+4.5\% / +5.6\% 
\\
Click & 
+13\% / +6.7\% &
+1.5\% / +0.5\% &
+2.7\% / +0.9\% 
\\
Appetency & 
+24\% / +0.7\% &
+1.6\% / -0.5\% &
+8.5\% / +0.7\%
\\
Churn & 
+12\% / +2.1\% &
+0.9\% / +1.3\% &
+1.6\% / +1.8\%
\\
Internet & 
+33\% / +22\% &
+2.6\% / +1.8\% &
+27\% / +19\%
\\
Upselling & 
+57\% / +50\% &
+1.6\% / +0.9\% &
+3.9\% / +2.9\% 
\\
Kick & 
+22\% / +28\% &
+1.3\% / +0.32\% &
+3.7\% / +3.3\% 
\\
\bottomrule
\end{tabular}
\end{minipage}
\hspace{0.3cm}
\begin{minipage}{0.47\textwidth}
\centering
\includegraphics[width=0.9\textwidth]{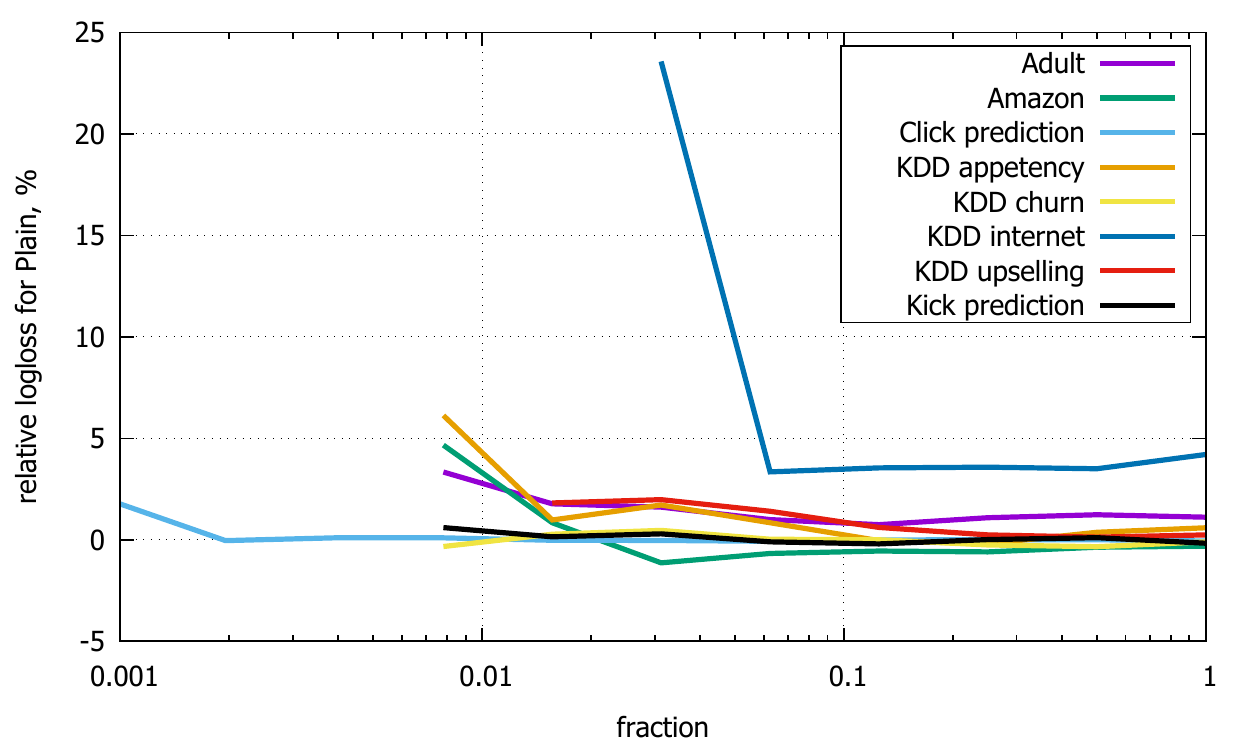}
\captionof{figure}{Relative error of Plain boosting mode compared to Ordered boosting mode depending on the fraction of the dataset.}\label{fig:plain_filtered}
\end{minipage}

\paragraph{Analysis of target statistics}

We compare different TSs introduced in Section~\ref{sec:TS} as options of CatBoost in Ordered boosting mode keeping all other algorithmic details the same; the results can be found in Table~\ref{tab:ctr}. Here, to save space, we present only relative increase in loss functions for each algorithm compared to CatBoost with ordered TS. Note that the ordered TS used in CatBoost significantly outperform all other approaches. Also, among the baselines, the holdout TS is the best for most of the datasets since it does not suffer from conditional shift discussed in Section~\ref{sec:TS} (P1); still, it is worse than CatBoost due to less effective usage of training data (P2). Leave-one-out is usually better than the greedy TS, but it can be much worse on some datasets, e.g., on Adult. The reason is that the greedy TS suffer from low-frequency categories, while the leave-one-out TS suffer also from high-frequency ones, and on Adult all the features have high frequency.

Finally, let us note that in Table~\ref{tab:ctr} we combine Ordered mode of CatBoost with different TSs. To generalize these results, we also made a similar experiment by combining different TS with Plain mode, used in standard gradient boosting. The obtained results and conclusions turned out to be very similar to the ones discussed above.

\paragraph{Feature combinations} 

The effect of feature combinations discussed in Section~\ref{sec:implementation} is demonstrated in Figure~\ref{fig:combinations} in Appendix~\ref{appendix::experimental_res}. In average, changing the number $c_{max}$ of features allowed to be combined from 1 to 2 provides an outstanding improvement of logloss by $1.86\%$ (reaching $11.3\%$), changing from 1 to 3 yields $2.04\%$, and further increase of $c_{max}$ does not influence the performance significantly.

\paragraph{Number of permutations}\label{sec:n_permut}

The effect of the number $s$ of permutations on the performance of CatBoost is presented in Figure~\ref{fig:folds} in Appendix~\ref{appendix::experimental_res}.  In average, increasing $s$ slightly decreases logloss, e.g., by $0.19\%$ for $s=3$ and by $0.38\%$ for $s=9$ compared to $s=1$. 

\section{Conclusion}

In this paper, we identify and analyze the problem of prediction shifts present in all existing implementations of gradient boosting. We propose a general solution, ordered boosting with ordered TS, which solves the problem. This idea is implemented in CatBoost, which is a new gradient boosting library. Empirical results demonstrate that CatBoost outperforms leading GBDT packages and leads to new state-of-the-art results on common benchmarks.

\subsubsection*{Acknowledgments}
We are very grateful to Mikhail Bilenko for important references and advices that lead to theoretical analysis of this paper, as well as suggestions on the presentation. We also thank Pavel Serdyukov for many helpful discussions and valuable links, Nikita Kazeev, Nikita Dmitriev, Stanislav Kirillov and Victor Omelyanenko for help with experiments.


\begin{appendices}

\section{Proof of Theorem 1}\label{appendix::proof}

\subsection{Proof for the case $\D_1 = \D_2$}

Let us denote by $A$ the event that each leaf in both stumps $h^1$ and $h^2$ contains at least one example, i.e., there exists at least one $\x \in \D$ with $\x^i = s$ for all $i \in \{1,2\}$, $s \in \{0,1\}$. All further reasonings are given conditioning on $A$. Note that the probability of $A$ is $1-O\left(2^{-n}\right)$, therefore we can assign an arbitrary value to any empty leaf during the learning process, and the choice of the value will affect all expectations we calculate below by $O\left(2^{-n}\right)$.

Denote by $\xi_{st}$, $s,t\in \{0,1\}$, the number of examples $\x_k\in\D$ with $\x_k=(s,t)$. The value of the first stump $h^1$ in the region $\{x^1=s\}$ is the average value of $y_k$ over examples from $\D$ belonging to this region. That is,
\begin{equation*}
h^1(0,t) = \frac{\sum_{j=1}^n c_2 \1_{\{x_j=(0,1)\}}}{\sum_{j=1}^n \1_{\{x_j^1=0\}} } = \frac{c_2 \xi_{01}}{\xi_{00}+\xi_{01}}\,,
\end{equation*}
\begin{equation*}
h^1(1,t) = \frac{\sum_{j=1}^n c_1\1_{\{x_j^1=1\}} + c_2 \1_{\{x_j=(1,1)\}}}{\sum_{j=1}^n \1_{\{x_j^1=1\}}} = c_1 + \frac{c_2 \xi_{11}}{\xi_{10}+\xi_{11}}\,.
\end{equation*}
Summarizing, we obtain 
\begin{equation}\label{eq:h1}
h^1(s,t)=c_1s + \frac{c_2\xi_{s 1}}{\xi_{s 0}+\xi_{s 1}}.
\end{equation}
Note that, by conditioning on $A$, we guarantee that the denominator $\xi_{s 0}+\xi_{s 1}$ is not equal to zero.

Now we derive the expectation $\E(h^1(\x))$ of prediction $h^1$ for a test example $\x=(s,t)$. 

It is easy to show that 
$\E\left(\frac{\xi_{s 1}}{\xi_{s 0}+\xi_{s 1}}\mid A\right) = \frac{1}{2}$. Indeed, due to the symmetry we have $\E\left(\frac{\xi_{s 1}}{\xi_{s 0}+\xi_{s 1}} \mid A\right) = \E\left(\frac{\xi_{s 0}}{\xi_{s 0}+\xi_{s 1}} \mid A\right)$ and the sum of these expectations is $\E\left(\frac{\xi_{s 0}+\xi_{s 1}}{\xi_{s 0}+\xi_{s 1}} \mid A\right) = 1$. So, by taking the expectation of~\eqref{eq:h1}, we obtain the following proposition.

\begin{prop}
We have $\E(h^1(s,t)\mid A) = c_1 s + \frac{c_2}{2}$.
\end{prop}

It means that the conditional expectation $\E(h^1(\x)\mid \x=(s,t), A)$ on a test example $\x$ equals $c_1s+\frac{c_2}{2}$, since $\x$ and $h^1$ are independent. 

\paragraph{Prediction shift of $h^1$} 
In this paragraph, we show that the conditional expectation $\E(h^1(\x_l)\mid \x_l=(s,t),A)$ on a training example $\x_l$ is shifted for any $l=1,\ldots, n$, because the model $h^1$ is fitted to $\x_l$. This is an auxiliary result, which is not used directly for proving the theorem, but helps to track the chain of obtained shifts.

\begin{prop}\label{prop2}
The conditional expectation is 
$$
\E(h^1(\x_l)\mid \x_l=(s,t),A) = c_1s+\frac{c_2}{2} - c_2\left(\frac{2t-1}{n}\right) + O(2^{-n})\,.
$$
\end{prop}

\begin{myprf}
Let us introduce the following notation 
$$
\alpha_{sk}=\frac{\1_{\{x_k=(s,1)\}}}{\xi_{s0}+\xi_{s1}}\,.
$$
Then, we can rewrite the conditional expectation as 
$$
c_1 s + c_2 \sum_{k=1}^n \E(\alpha_{sk}\mid \x_l=(s,t), A)\,.
$$ 
Lemma~\ref{lem1} below implies that
$\E(\alpha_{sl}\mid \x_l=(s,t),A) = \frac{2t}{n}$.
For $k\neq l$, we have 
\begin{multline*}
\E(\alpha_{sk}\mid \x_l=(s,t),A) = \frac{1}{4}\, \E\left(\frac{1}{\xi_{s0}+\xi_{s1}}\mid \x_l=(s,t), \x_k=(s,1), A\right) \\ = \frac{1}{2n} \left(1-\frac{1}{n-1}+\frac{n-2}{\left(2^{n-1}-2\right)(n-1)}\right)
\end{multline*}
due to Lemma~\ref{lem2} below. Finally, we obtain 
\begin{multline*}
\E(h^1(\x_l)\mid \x_l=(s,t)) = c_1s + c_2 \left( \frac{2t}{n} + (n-1) \frac{1}{2n} \left(1-\frac{1}{n-1}\right)\right)\\ +O\left(2^{-n}\right)  = c_1s + \frac{c_2}{2} - c_2\left(\frac{2t-1}{n}\right)+O(2^{-n}).
\end{multline*}
\end{myprf}

\begin{lem}\label{lem1}
$\E\left(\frac{1}{\xi_{s0}+\xi_{s1}}\mid \x_1=(s,t),A\right) = \frac{2}{n}$\,.
\end{lem}
\begin{myprf}
Note that given $\x_1 = (s,t)$, $A$ corresponds to the event that there is an example with $x^1 = 1-s$ and (possibly another) example with $x^2 = 1-t$ among $\x_2, \ldots, \x_n$.

Note that $\xi_{s0}+\xi_{s1} = \sum_{j=1}^n \1_{\{x_j^1=s\}}$.
For $k = 1, \ldots, n-1$, we have
$$
\Prob(\xi_{s0}+\xi_{s1}= k\mid \x_1=(s,t),A) = 
\frac{\Prob(\xi_{s0}+\xi_{s1}= k, A \mid \x_1=(s,t))}{\Prob(A\mid \x_1=(s,t))} 
= \frac{\binom{n-1}{k-1}}{2^{n-1}\left(1 - 2^{-(n-1)}\right)},
$$
since $\1_{\{x_1^1=s\}}=1$ when $\x_1=(s,t)$ with probability 1, $\sum_{j=2}^n \1_{\{x_j^1=s\}}$ is a binomial variable independent of $\x_1$, and an example with $x^1 = 1-s$ exists whenever $\xi_{s0}+\xi_{s1}= k<n$ and $\x_1=(s,t)$ (while the existence of one with $x^2 = 1-t$ is an independent event). Therefore, we have 
$$
\E\left(\frac{1}{\xi_{s0}+\xi_{s1}}\mid \x_1=(s,t),A\right) 
= \sum_{k=1}^{n-1} \frac{1}{k} \frac{\binom{n-1}{k-1}}{2^{n-1}-1} 
= \frac{1}{n \left(2^{n-1}-1\right)} \sum_{k=1}^{n-1} \binom{n}{k} = \frac{2}{n}\,.
$$
\end{myprf}

\begin{lem}\label{lem2}
We have
$$
\E\left(\frac{1}{\xi_{s0}+\xi_{s1}}\mid \x_1=(s,t_1),\x_2=(s,t_2),A\right) =  \frac{2}{n}\left(1-\frac{1}{n-1}+\frac{n-2}{\left(2^{n-1}-2\right)(n-1)}\right).
$$
\end{lem}

\begin{myprf}
Similarly to the previous proof, for $k = 2, \ldots, n-1$, we have 
$$
\Prob\left(\xi_{s0}+\xi_{s1} = k\mid \x_1=(s,t_1),\x_2=(s,t_2),A\right) 
= \frac{\binom{n-2}{k-2}}{2^{n-2}\left(1 - 2^{-(n-2)}\right)}\,.
$$
Therefore,
\begin{multline*}
\E\left(\frac{1}{\xi_{s0}+\xi_{s1}} \mid \x_1=(s,t_1),\x_2=(s,t_2),A\right)  =\frac{1}{2^{n-2}\left(1 - 2^{-(n-1)}\right)} 
\sum_{k=2}^{n-1} \frac{\binom{n-2}{k-2}}{k} \\
 = \frac{1}{2^{n-2}-1} 
 \sum_{k=2}^{n-1} \binom{n-2}{k-2} \left ( \frac{1}{k-1}-\frac{1}{(k-1)k} \right ) \\
= \frac{1}{2^{n-2}-1}
\sum_{k=2}^{n-1}\left( \frac{1}{n-1} \binom{n-1}{k-1} - \frac{1}{n(n-1)} \binom{n}{k} \right ) =\\
= \frac{1}{2^{n-2}-1}
\left( \frac{1}{n-1} (2^{n-1}-2) - \frac{1}{n(n-1)}(2^n-n-2) \right ) =\\
= \frac{2}{n}\left(1-\frac{1}{n-1}+\frac{n-2}{\left(2^{n-1}-2\right)(n-1)}\right)\,.
\end{multline*}
\end{myprf}

\paragraph{Bias of the model $h^1+h^2$}
Proposition~\ref{prop2} shows that the values of the model $h^1$ on training examples are shifted with respect to the ones on test examples. The next step is to show how this can lead to a bias of the trained model, if we use the same dataset for building both $h^1$ and $h^2$. Namely, we derive the expected value of $h^1(s,t)+h^2(s,t)$ and obtain a bias according to the following result.

\begin{prop}\label{prop3}
If both $h^1$ and $h^2$ are built using the same dataset $\D$, then
$$
\E \left(h^1(s,t)+h^2(s,t)\mid A\right)= f^*(s,t)-\frac{1}{n-1}c_2\left(t-\frac{1}{2}\right)+O(1/2^{n})\,.
$$
\end{prop}

\begin{myprf}
The residual after the first step is 
$$
f^*(s,t)-h^1(s,t) = c_2\left(t-\frac{\xi_{s1}}{\xi_{s0}+\xi_{s1}}\right)\,.
$$
Therefore, we get
$$
h^2(s,t) = \frac{c_2}{\xi_{0t}+\xi_{1t}}\left(\left(t-\frac{\xi_{01}}{\xi_{00}+\xi_{01}}\right)\xi_{0t} + \left(t-\frac{\xi_{11}}{\xi_{10}+\xi_{11}}\right)\xi_{1t}\right)\,,
$$
which is equal to 
$$
-c_2\left (\frac{\xi_{00}\xi_{01}}{(\xi_{00}+\xi_{01})(\xi_{00}+\xi_{10})} + \frac{\xi_{10}\xi_{11}}{(\xi_{10}+\xi_{11})(\xi_{00}+\xi_{10})}\right )
$$ 
for $t=0$ and to 
$$
c_2\left (   \frac{\xi_{00}\xi_{01}}{(\xi_{00}+\xi_{01})(\xi_{01}+\xi_{11})} + \frac{\xi_{10}\xi_{11}}{(\xi_{10}+\xi_{11})(\xi_{01}+\xi_{11})} \right )
$$ 
for $t=1$. The expected values of all four ratios are equal due to symmetries, and they are equal to $\frac{1}{4}\left(1-\frac{1}{n-1}\right)+O(2^{-n})$ according to Lemma~\ref{lem3} below. 
So, we obtain 
$$
\E(h^2(s,t)\mid A) = (2t-1)\frac{c_2}{2}\left(1-\frac{1}{n-1}\right)+O(2^{-n})
$$ 
and 
$$
\E(h^1(s,t)+h^2(s,t)\mid A)= f^*(s,t) - c_2\frac{1}{n-1}\left(t-\frac{1}{2}\right)+O(2^{-n})\,.
$$
\end{myprf}

\begin{lem}\label{lem3}
We have 
$$
\E\left(\frac{\xi_{00}\xi_{01}}{(\xi_{00}+\xi_{01})(\xi_{01}+\xi_{11})}\mid A\right) = \frac{1}{4}\left(1-\frac{1}{n-1}\right)+O(2^{-n})\,.
$$
\end{lem}

\begin{myprf}
First, linearity implies 
$$
\E\left(\frac{\xi_{00}\xi_{01}}{(\xi_{00}+\xi_{01})(\xi_{01}+\xi_{11})}\mid A\right)  = \sum_{i,j} \E\left(\frac{\1_{\x_i=(0,0),\x_j=(0,1)}}{(\xi_{00}+\xi_{01})(\xi_{01}+\xi_{11})}\mid A\right)\,.
$$
Taking into account that all terms are equal, the expectation can be written as $\frac{n(n-1)}{4^2}  a$, where
$$
a=\E\left(\frac{1}{(\xi_{00}+\xi_{01})(\xi_{01}+\xi_{11})}\mid \x_1=(0,0), \x_2=(0,1), A\right)\,.
$$
A key observation is that $\xi_{00}+\xi_{01}$ and $\xi_{01}+\xi_{11}$ are two independent binomial variables: the former one is the number of $k$ such that ${x_k^1=0}$ and the latter one is the number of $k$ such that $x_k^2=1$.  Moreover, they (and also their inverses) are also conditionally independent given that first two observations of the Bernoulli scheme are known ($\x_1=(0,0), \x_2=(0,1)$) and given $A$. This conditional independence implies that $a$ is the product of $\E\left(\frac{1}{\xi_{00}+\xi_{01}}\mid \x_1=(0,0), \x_2=(0,1), A\right)$ 
and 
$\E\left(\frac{1}{\xi_{01}+\xi_{11}}\mid \x_1=(0,0), \x_2=(0,1), A\right)$. 
The first factor equals 
$\frac{2}{n}\left(1-\frac{1}{n-1}+O (2^{-n})\right)$ 
according to Lemma~\ref{lem2}. The second one is equal to
$
\E\left(\frac{1}{\xi_{01}+\xi_{11}}\mid \x_1=(0,0), \x_2=(0,1)\right)
$
since $A$ does not bring any new information about the number of $k$ with $x_k^2 = 1$ given $\x_1=(0,0), \x_2=(0,1)$. So, according to Lemma~\ref{lem4} below, the second factor equals $\frac{2}{n-1}(1+O(2^{-n}))$. Finally, we obtain 
\begin{multline*}
\E\left(\frac{\xi_{00}\xi_{01}}{(\xi_{00}+\xi_{01})(\xi_{01}+\xi_{11})}\right)
\\ = \frac{n(n-1)}{4^2}\frac{4}{n(n-1)}\left(1-\frac{1}{n-1}\right)+O(2^{-n})=\frac{1}{4}\left(1-\frac{1}{n-1}\right)+O(2^{-n}).
\end{multline*}
\end{myprf}

\begin{lem}\label{lem4}
$\E\left(\frac{1}{\xi_{01}+\xi_{11}}\mid \x_1=(0,0), \x_2=(0,1)\right) = \frac{2}{n-1} - \frac{1}{2^{n-2}(n-1)}$\,.
\end{lem}

\begin{myprf}
Similarly to the proof of Lemma~\ref{prop2}, we have
$$
\Prob(\xi_{01}+\xi_{11}= k\mid \x_1=(0,0), \x_2=(0,1) )
= \binom{n-2}{k-1}2^{-(n-2)}\,.
$$
Therefore, we get 
\begin{multline*}
\E\left(\frac{1}{\xi_{01}+\xi_{11}}\mid \x_1=(0,0), \x_2=(0,1)\right)
= \sum_{k=1}^{n-1} \frac{1}{k} \binom{n-2}{k-1}2^{-(n-2)} \\
=  \frac{2^{-(n-2)}}{n-1}\sum_{k=1}^{n-1} \binom{n-1}{k} = \frac{2}{n-1} - \frac{1}{2^{n-2}(n-1)}\,.
\end{multline*}
\end{myprf}

\subsection{Proof for independently sampled $\D_1$ and $\D_2$}

Assume that we have an additional sample $\D_2 = \{\x_{n+k}\}_{k=1..n}$ for building $h^2$. Now $A$ denotes the event that each leaf in $h^1$ contains at least one example from $\D_1$ and each leaf in $h^2$ contains at least one example from $\D_2$.

\begin{prop}
If $h^2$ is built using dataset $\D_2$, then
$$
\E( h^1(s,t)+h^2(s,t) \mid A) = f^*(s,t)\,.
$$ 
\end{prop}

\begin{myprf}

Let us denote by $\xi'_{st}$ the number of examples $\x_{n+k}$ that are equal to $(s,t)$, $k=1,\ldots, n$.

First, we need  to derive the expectation $\E(h^2(s,t))$ of $h^2$ on a test example $\x=(s,t)$. Similarly to the proof of Proposition~\ref{prop3}, we get
$$
h^2(s,0) = -c_2\left (\frac{\xi'_{00}\xi_{01}}{(\xi_{00}+\xi_{01})(\xi'_{00}+\xi'_{10})} + \frac{\xi'_{10}\xi_{11}}{(\xi_{10}+\xi_{11})(\xi'_{00}+\xi'_{10})}\right)\,,
$$ 
$$
h^2(s,1) = c_2\left (   \frac{\xi_{00}\xi'_{01}}{(\xi_{00}+\xi_{01})(\xi'_{01}+\xi'_{11})} + \frac{\xi_{10}\xi'_{11}}{(\xi_{10}+\xi_{11})(\xi'_{01}+\xi'_{11})} \right)\,.
$$

Due to the symmetries, the expected values of all four fractions above are equal. Also, due to the independence of $\xi_{ij}$ and $\xi'_{kl}$, we have
$$
\E \left(\frac{\xi'_{00}\xi_{01}}{(\xi_{00}+\xi_{01})(\xi'_{00}+\xi'_{10})} \mid A\right) 
= \E \left(\frac{\xi_{01}}{\xi_{00}+\xi_{01}} \mid A \right) \E \left( \frac{\xi'_{00}}{\xi'_{00}+\xi'_{10}} \mid A\right) = \frac{1}{4}\,.
$$
Therefore, $\E(h^2(s,0) \mid A) = -\frac{c_2}{2}$ and  $\E(h^2(s,1)\mid A) = \frac{c_2}{2}$.

Summing up, $\E(h^2(s,t)\mid A)= c_2t-\frac{c_2}{2}$ and $\E(h^1(s,t)+h^2(s,t)\mid A)= c_1s+c_2t$.
\end{myprf}

\section{Formal description of CatBoost algorithm}\label{appendix::full_description}

In this section, we formally describe the CatBoost algorithm introduced in Section~\ref{sec:implementation}. In Algorithm~\ref{alg:catboost}, we provide more information on particular details including the speeding up trick introduced in paragraph ``Complexity''. The key step of the CatBoost algorithm is the procedure of building a tree described in detail in Function~\ref{alg:CatBoostStep}. To obtain the formal description of the CatBoost algorithm without the speeding up trick, one should replace $\ceil{\log_2 n}$ by $n$ in line~6 of Algorithm~\ref{alg:catboost} and use Algorithm~\ref{alg:CatBoostStepMain} instead of Function~\ref{alg:CatBoostStep}.

We use Function $GetLeaf(\x,T,\sigma_r)$ to describe how examples are matched to leaves $leaf_r(i)$. Given an example with features $\x$, we calculate ordered TS on the basis of the permutation $\sigma_r$ and then choose the leaf of tree $T$ corresponding to features $\x$ enriched by the obtained ordered TS. Using $ApplyMode$ instead of a permutation in function $GetLeaf$ in line~15 of Algorithm~\ref{alg:catboost} means that we use TS calculated on the whole training data to apply the trained model on a new example.

\begin{algorithm}
\SetKwInOut{Input}{input}\SetKwInOut{Output}{output}
\Input{\,\,$\{(\x_i,y_i)\}_{i=1}^n$, $I$, $\alpha$, $L$, $s$, $Mode$}
  \BlankLine
  \ShowLn $\sigma_r \leftarrow$ random permutation of $[1,n]$ for $r = 0..s$\;
  \ShowLn $M_{0}(i) \leftarrow 0$ for $i = 1..n$\;
  \ShowLn \If{$Mode = Plain$}{
    \ShowLn $M_{r}(i) \leftarrow 0$ for $r = 1..s$, $i: \sigma_r(i) \le 2^{j+1}$\;}
  \ShowLn \If{$Mode=Ordered$}{
    \ShowLn \For{$j\leftarrow 1$ \KwTo $\ceil{\log_2 n}$}{
    \ShowLn $M_{r,j}(i) \leftarrow 0$ for $r = 1..s$, 
    $i = 1..2^{j+1}$\;}
  } 
  \ShowLn \For{$t$ $\leftarrow 1$ \KwTo $I$}{ 
      \ShowLn $T_t,\ \{M_r\}_{r=1}^s \leftarrow BuildTree(\{M_r\}_{r=1}^s,\{(\x_i,y_i)\}_{i=1}^n,\alpha,L,\{\sigma_i\}_{i=1}^s, Mode)$\;
      \ShowLn  $leaf_0(i) \leftarrow GetLeaf(\x_i,T_t, \sigma_0)$ for $i=1..n$\;
     \ShowLn  $grad_0 \leftarrow CalcGradient(L,M_0,y)$\;
     \ShowLn  \ForEach{leaf $j$ in $T_t$}{
      \ShowLn     $b_j^{t} \leftarrow -\mathrm{avg}(grad_{0}(i)$ for $i:\ leaf_0(i)=j)$\;
          }
     \ShowLn  $M_{0}(i) \leftarrow M_{0}(i) + \alpha b_{leaf_{0}(i)}^t$ for $i=1..n$\;
  }
 \ShowLn \Return $F(\x) = \sum_{t = 1}^I \sum_j \alpha \, b_j^{t} \1_{\{GetLeaf(\x,T_t,ApplyMode) =j\}}$\;
  \caption{CatBoost}
  \label{alg:catboost}
\end{algorithm}

\begin{function}
  \SetKwInOut{Input}{input}\SetKwInOut{Output}{output}
  \Input{\,\,$M$,$\{(\x_i,y_i)\}_{i=1}^n$, $\alpha$, $L$, $\{\sigma_i\}_{i=1}^s$, $Mode$ }
  \BlankLine 
  \ShowLn $grad \leftarrow CalcGradient(L,M,y)$\;
  \ShowLn $r \leftarrow random(1,s)$\;
  \ShowLn \If{$Mode = Plain$}{
     \ShowLn $G \leftarrow (grad_{r}(i) \mbox{ for } i=1..n)$\;}
  \ShowLn \If{$Mode=Ordered$}{
    \ShowLn $G \leftarrow (grad_{r,\floor{\log_2 (\sigma_r(i)-1)}}(i) \mbox{ for } i=1..n)$\;}
  \ShowLn $T \leftarrow$ empty tree\;
  \ShowLn \ForEach{step of top-down procedure}{
   \ShowLn \ForEach{candidate split $c$ }{
      \ShowLn $T_c \leftarrow$ add split $c$ to $T$\;
      \ShowLn $leaf_r(i) \leftarrow GetLeaf(\x_i,T_c, \sigma_r)$ for $i=1..n$\;
      \ShowLn \If{$Mode = Plain$}{
              \ShowLn $\Delta(i) \leftarrow \mathrm{avg}(grad_{r}(p)$ for $p:\ leaf_r(p)=leaf_r(i) )$ \ for $i=1..n$\; }
      \ShowLn \If{$Mode=Ordered$}{
              \ShowLn $\Delta(i) \leftarrow \mathrm{avg}(grad_{r,\floor{\log_2 (\sigma_r(i)-1)}}(p)$ for $p:\  leaf_r(p)=leaf_r(i), \sigma_r(p)<\sigma_r(i))$ \ for $i=1..n$\;}
       \ShowLn $loss(T_c) \leftarrow \cos(\Delta,G)$
  }
   \ShowLn $T \leftarrow \argmin_{T_c}(loss(T_c))$
   }
   \ShowLn $leaf_{r'}(i) \leftarrow GetLeaf(\x_i,T, \sigma_{r'})$ for $r'=1..s$, $i=1..n$\;
  \ShowLn \If{$Mode = Plain$}{
     \ShowLn $M_{r'}(i) \leftarrow M_{r'}(i) - \alpha \, \mathrm{avg}(grad_{r'}(p)$ for $p:\ leaf_{r'}(p)=leaf_{r'}(i) )$ for $r'=1..s$, $i=1..n$\; }
  \ShowLn \If{$Mode=Ordered$}{
    \ShowLn \For{$j\leftarrow 1$ \KwTo $\ceil{\log_2 n}$}{
    \ShowLn $M_{r',j}(i) \leftarrow M_{r',j}(i) - \alpha \, \mathrm{avg}(grad_{r',j}(p)$ for $p:\  leaf_{r'}(p)=leaf_{r'}(i), \sigma_{r'}(p)\leq 2^j)$ for $r'=1..s$,\ \  
    $i: \sigma_{r'}(i) \le 2^{j+1}$\;}
} 
  \ShowLn \Return{$T, M$}
  {\caption{$BuildTree$()}
      \label{alg:CatBoostStep}}
\end{function} 

\section{Time complexity analysis}\label{appendix::time_complexity}

\subsection{Theoretical analysis}\label{appendix::time_complexity_theory}

We present the computational complexity of different components of any of the two modes of CatBoost per one iteration in Table~\ref{tab:complexity_supplementary}.

\begin{table}[h]
\caption{Computational complexity.}
\label{tab:complexity_supplementary}
\centering
\begin{tabular}{lccccc}
  \toprule
 Procedure & CalcGradient & Build $T$ & Calc values $b^t_j$ & Update $M$ & Calc ordered TS  \\
  \midrule
 \makecell{Complexity\\for iteration $t$} &  $O(s\cdot n)$ &  $O(|C|\cdot n)$ & $O(n)$  &   $O(s\cdot n)$ & $O(N_{TS,t}\cdot n)$\\
  \bottomrule
\end{tabular}
\end{table}

We first prove these asymptotics for the Ordered mode. For this purpose, we estimate the number $N_{pred}$ of predictions $M_{r,j}(i)$ to be maintained: 
$$
N_{pred} = (s+1)\cdot  \sum_{j=1}^{\ceil{\log_2 n}} 2^{j+1} < (s+1)\cdot 2^{\log_2 n + 3}=8(s+1) n\,.
$$
Then, obviously, the complexity of CalcGradient is $O(N_{pred}) = O(s\cdot n)$. The complexity of leaf values calculation is $O(n)$, since each example $i$ is included only in averaging operation in leaf $leaf_0(i)$. 

Calculation of the ordered TS for one categorical feature can be performed sequentially in the order of the permutation by $n$ additive operations for calculation of $n$ partial sums and $n$ division operations. Thus, the overall complexity of the procedure is $O(N_{TS,t}\cdot n)$, where $N_{TS,t}$ is the number of TS which were not calculated on the previous iterations. Since the leaf values $\Delta(i)$ calculated in line~15 of Function~\ref{alg:CatBoostStep} can be considered as ordered TS, where gradients play the role of targets, the complexity of building a tree $T$ is $O(|C|\cdot n)$, where $C$ is the set of candidate splits to be considered at the given iteration.
Finally, for updating the supporting models (lines 22-23 in Function~\ref{alg:CatBoostStep}), we need to perform one averaging operation for each $j=1,\ldots,\ceil{\log_2 n}$, and each maintained gradient $grad_{r',j}(p)$ is included in one averaging operation. Thus, the number of operations is bounded by the number of the maintained gradients $grad_{r',j}(p)$, which is equal to $N_{pred}=O(s\cdot n)$. 

To finish the proof, note that any component of the Plain mode is not less efficient than the same one of the Ordered mode but, at the same time, cannot be more efficient than corresponding asymptotics from Table~\ref{tab:complexity_supplementary}.

\subsection{Empirical analysis}\label{appendix::time_complexity_experiments}

It is quite hard to compare different boosting libraries in terms of training speed. Every algorithm has a vast number of parameters which affect training speed, quality and model size in a non-obvious way. Different libraries have their unique quality/training speed trade-off's and they cannot be compared without domain knowledge (e.g., is $0.5\%$ of quality metric worth it to train a model 3-4 times slower?). Plus for each library it is possible to obtain almost the same quality with different ensemble sizes and parameters. As a result, one cannot compare libraries by time needed to obtain a certain level of quality. As a result, we could give only some insights of how fast our implementation could train a model of a fixed size. We use Epsilon dataset and we measure mean tree construction time one can achieve without using feature subsampling and/or bagging by CatBoost (both Ordered and Plain modes), XGBoost (we use histogram-based version, which is faster) and LightGBM.  For XGBoost and CatBoost we use the default tree depth equal to 6, for LightGBM we set leaves count to 64 to have comparable results. 
We run all experiments on the same machine with Intel Xeon E3-12xx 2.6GHz, 16 cores, 64GB RAM and run all algorithms with 16 threads. 

We set such learning rate that algorithms start to overfit approximately after constructing about 7000 trees and measure the average time to train ensembles of 8000 trees. Mean tree construction time is presented in Table~\ref{tab:speed}. Note that CatBoost Plain and LightGBM are the fastest ones followed by Ordered mode, which is about 1.7 times slower, which is expected. 

\begin{table}
\caption{Comparison of running times on Epsilon}\label{tab:speed}
\centering
\begin{tabular}{lc}
\cmidrule{2-2}
&  time per tree\\
\midrule
CatBoost Plain 
& \bf 1.1 s  \\
CatBoost Ordered 
& 1.9 s  \\
XGBoost 
& 3.9 s \\
LightGBM 
& \bf 1.1 s  \\
\bottomrule
\end{tabular}
\end{table}

Finally, let us note that CatBoost has a highly efficient GPU implementation. The detailed description and comparison of the running times are beyond the scope of the current article, but these experiments can be found on the corresponding GitHub page.\footnote{\url{https://github.com/catboost/benchmarks/tree/master/gpu_training}}

\section{Experimental setup}\label{appendix::experimental_setup}

\subsection{Description of the datasets}

The datasets used in our experiments are described in Table~\ref{tab:datasets}. 

\begin{table}
\caption{Description of the datasets.}\label{tab:datasets}
\begin{center}
    \begin{tabular}{ | l | c | c | p{7cm}|}
    \hline
    Dataset name &  Instances & Features & Description \\ \hline
    \hline   Adult\protect\footnotemark & 48842 & 15 & Prediction task is to determine whether a person makes over 50K a year. Extraction was done by Barry Becker from the 1994 Census database. A set of reasonably clean records was extracted using the following conditions: (AAGE>16) and (AGI>100) and (AFNLWGT>1) and (HRSWK>0)\\\hline

Amazon\protect\footnotemark & 32769 & 10 & Data from the Kaggle Amazon Employee challenge.\\\hline
   
    Click Prediction\protect\footnotemark & 399482 & 12 & This data is derived from the 2012 KDD Cup. The data is subsampled to 1\% of the original number of instances, downsampling the majority class (click=0) so that the target feature is reasonably balanced (5 to 1). The data is about advertisements shown alongside search results in a search engine, and whether or not people clicked on these ads. The task is to build the best possible model to predict whether a user will click on a given ad.\\\hline 
    
Epsilon\protect\footnotemark & 400000  & 2000  & PASCAL Challenge 2008. \\\hline
 
KDD appetency\protect\footnotemark & 50000 & 231 & Small version of KDD 2009 Cup data.\\\hline

KDD churn\protect\footnotemark & 50000 & 231 & Small version of KDD 2009 Cup data.\\\hline
   
KDD Internet\protect\footnotemark & 10108 & 69 & Binarized version of the original dataset. The multi-class target feature is converted to a two-class nominal target feature by re-labeling the majority class as positive (`P') and all others as negative (`N'). Originally converted by Quan Sun.\\\hline

KDD upselling\footnotemark & 50000 & 231 & Small version of KDD 2009 Cup data.\\\hline
        
Kick prediction\protect\footnotemark & 72983 & 36 & Data from ``Don't Get Kicked!''\ Kaggle challenge.\\\hline
\end{tabular}
\end{center}
\end{table}

\addtocounter{footnote}{-8}
\footnotetext{\url{https://archive.ics.uci.edu/ml/datasets/Adult}}
\addtocounter{footnote}{1}
\footnotetext{\url{https://www.kaggle.com/c/amazon-employee-access-challenge}}
\addtocounter{footnote}{1}
\footnotetext{\url{http://www.kdd.org/kdd-cup/view/kdd-cup-2012-track-2}}
\addtocounter{footnote}{1}
\footnotetext{\url{https://www.csie.ntu.edu.tw/~cjlin/libsvmtools/datasets/binary.html}}
\addtocounter{footnote}{1}
\footnotetext{\url{http://www.kdd.org/kdd-cup/view/kdd-cup-2009/Data}}
\addtocounter{footnote}{1}
\footnotetext{\url{http://www.kdd.org/kdd-cup/view/kdd-cup-2009/Data}}
\addtocounter{footnote}{1}
\footnotetext{{\url{https://kdd.ics.uci.edu/databases/internet\_usage/internet\_usage.html}}}
\addtocounter{footnote}{1}
\footnotetext{\url{http://www.kdd.org/kdd-cup/view/kdd-cup-2009/Data}} 
\addtocounter{footnote}{1}
\footnotetext{\url{https://www.kaggle.com/c/DontGetKicked}}

\subsection{Experimental settings}\label{sec:ExerimentalSettings}
In our experiments, we evaluate different modifications of CatBoost and two popular gradient boosting libraries: LightGBM and XGBoost. All the code needed for reproducing our experiments is published on our GitHub\footnote{\url{https://github.com/catboost/benchmarks/tree/master/quality_benchmarks}}.

\paragraph{Train-test splits}
Each dataset was randomly split into training set (80\%) and test set (20\%). We denote them as $D_{full\_train}$ and $D_{test}$.

We use 5-fold cross-validation to tune parameters of each model on the training set. Accordingly, $D_{full\_train}$ is randomly split into 5 equally sized parts $D_1, \dots, D_5$ (sampling is stratified by classes). These parts are used to construct 5 training and validation sets:  $D_i^{train} = \cup_{j\neq i}D_j$ and $D_i^{val} = D_i$ for $1 \le i \le 5$.

\paragraph{Preprocessing}
We applied the following steps to datasets with missing values:
\begin{itemize}
\item
For categorical variables, missing values are replaced with a special value, i.e., we treat missing values as a special category;

\item
For numerical variables, missing values are replaced with zeros, and a binary dummy feature for each imputed feature is added.
\end{itemize}

For XGBoost, LightGBM and the raw setting of CatBoost (see Appendix~\ref{appendix::experimental_res}), we perform the following preprocessing of categorical features. For each pair of datasets $(D^{train}_i,\ D^{val}_i)$, $i=1, \ldots, 5$, and $(D_{full\_train},D_{test})$, we preprocess the categorical features by calculating ordered TS (described in Section~\ref{sec:TS}) on the basis of a random permutation of the examples of the first (training) dataset. All the permutations are generated independently. The resulting values of TS are considered as numerical features by any algorithm to be evaluated.

\paragraph{Parameter Tuning}

We tune all the key parameters of each algorithm by 50 steps of the sequential optimization algorithm Tree Parzen Estimator implemented in Hyperopt library\footnote{\url{https://github.com/hyperopt/hyperopt}} (mode \textit{algo=tpe.suggest}) by minimizing logloss. Below is the list of the tuned parameters and their distributions the optimization algorithm started from:

\medskip
\noindent XGBoost:
\begin{itemize}
  \item `eta': Log-uniform distribution $[e^{-7}, 1]$
  \item `max\_depth': Discrete uniform distribution $[2, 10]$
  \item `subsample': Uniform $[0.5, 1]$
  \item `colsample\_bytree': Uniform $[0.5, 1]$
  \item `colsample\_bylevel': Uniform $[0.5, 1]$
  \item `min\_child\_weight': Log-uniform distribution $[e^{-16}, e^{5}]$
  \item `alpha': Mixed: $0.5\, \cdot$ Degenerate at 0 + $0.5\, \cdot$ Log-uniform distribution $[e^{-16}, e^{2}]$
  \item `lambda': Mixed: $0.5\, \cdot$ Degenerate at 0 + $0.5\, \cdot$ Log-uniform distribution $[e^{-16}, e^{2}]$
  \item `gamma': Mixed: $0.5\, \cdot$ Degenerate at 0 + $0.5\, \cdot$ Log-uniform distribution $[e^{-16}, e^{2}]$
\end{itemize}

\medskip
\noindent LightGBM:
\begin{itemize}
  \item `learning\_rate': Log-uniform distribution $[e^{-7}, 1]$
  \item `num\_leaves' : Discrete log-uniform distribution $[1, e^{7}]$
  \item `feature\_fraction': Uniform $[0.5, 1]$
  \item `bagging\_fraction': Uniform $[0.5, 1]$
  \item `min\_sum\_hessian\_in\_leaf': Log-uniform distribution $[e^{-16}, e^{5}]$
  \item `min\_data\_in\_leaf':  Discrete log-uniform distribution $[1, e^{6}]$
  \item `lambda\_l1': Mixed: $0.5\, \cdot$ Degenerate at 0 + $0.5\, \cdot$ Log-uniform distribution $[e^{-16}, e^{2}]$
  \item `lambda\_l2': Mixed: $0.5\, \cdot$ Degenerate at 0 + $0.5\, \cdot$ Log-uniform distribution $[e^{-16}, e^{2}]$
\end{itemize}

\medskip
\noindent CatBoost:
\begin{itemize}
  \item `learning\_rate':  Log-uniform distribution $[e^{-7}, 1]$
  \item `random\_strength': Discrete uniform distribution over a set  $\{1, 20\}$
  \item `one\_hot\_max\_size': Discrete uniform distribution over a set  $\{0, 25\}$
  \item `l2\_leaf\_reg': Log-uniform distribution $[1, 10]$
  \item `bagging\_temperature': Uniform $[0, 1]$
  \item `gradient\_iterations' : Discrete uniform distribution over a set  $\{1, 10\}$
\end{itemize}

Next, having fixed all other parameters, we perform exhaustive search for the number of trees in the interval $[1,5000]$. We collect logloss value for each training iteration from 1 to 5000 for each of the 5 folds. Then we choose the iteration with minimum logloss averaged over 5 folds.

For evaluation, each algorithm was run on the preprocessed training data $D_{full\_train}$ with the tuned parameters. The resulting model was evaluated on the preprocessed test set $D_{test}$.

\paragraph{Versions of the libraries}

\begin{itemize}
  \item catboost (0.3)
  \item xgboost (0.6)
  \item scikit-learn (0.18.1)
  \item scipy (0.19.0)
  \item pandas (0.19.2)
  \item numpy (1.12.1)
  \item lightgbm (0.1)
  \item hyperopt (0.0.2)
  \item h2o (3.10.4.6)
  \item R (3.3.3)
\end{itemize}

\section{Analysis of iterated bagging}\label{appendix::iterated_bagging}
Based on the out-of-bag estimation~\cite{breiman1996out}, Breiman proposed \textit{iterated bagging} \cite{breiman2001using} which simultaneously constructs $K$ models $F_i$, $i=1,\ldots, K$, associated with $K$ independently bootstrapped subsamples~$\D_i$. At $t$-th step of the process, models $F^{t}_i$ are grown from their predecessors $F^{t-1}_i$ as follows. The current estimate $M^t_j$ at example $j$ is obtained as the average of the outputs of all models $F^{t-1}_k$ such that $j \notin \D_k$. The term $h^{t}_i$ is built as a predictor of the residuals $r^t_j:=y_j-M^t_j$ (targets minus current estimates) on $\D_i$. Finally, the models are updated: $F^{t}_i:=F^{t-1}_i+h^{t}_i$. Unfortunately, the residuals $r^t_j$ used in this procedure are not unshifted (in terms of Section~\ref{sec:biasness_definition}), or unbiased (in terms of \textit{iterated bagging}), because each model $F^{t}_i$ depends on each observation $(\x_j, y_j)$ by construction. Indeed, although $h^t_k$ does not use $y_j$ directly, if $j\notin \D_k$, it still uses $M^{t-1}_{j'}$ for $j'\in \D_k$, which, in turn, can depend on $(\x_j,y_j)$.

Also note that computational complexity of this algorithm exceeds one of classic GBDT by factor of~$K$.

\section{Ordered boosting with categorical features}\label{appendix::combination}

In Sections~\ref{sec:TS} and~\ref{sec:ordered_boosting}, we proposed to use some random permutations $\sigma_{cat}$ and $\sigma_{boost}$ of training examples for the TS calculation and for ordered boosting, respectively. Now, being combined in one algorithm, should these two permutations be somehow dependent? We argue that they should coincide. Otherwise, there exist examples $\x_i$ and $\x_j$ such that $\sigma_{boost}(i)<\sigma_{boost}(j)$ and $\sigma_{cat}(i)>\sigma_{cat}(j)$. Then, the model $M_{\sigma_{boost}(j)}$ is trained using TS features of, in particular, example $\x_i$, which are calculated using $y_j$. In general, it may shift the prediction $M_{\sigma_{boost}(j)}(\x_j)$. To avoid such a shift, we set $\sigma_{cat}=\sigma_{boost}$ in CatBoost. In the case of the ordered boosting (Algorithm~\ref{alg:ordered}) with sliding window TS\footnote{Ordered TS calculated on the basis of a fixed number of preceding examples (both for training and test examples).}
it guarantees that the prediction $M_{\sigma(i)-1}(\x_i)$ is not shifted for $i=1,\ldots,n$, since, first, the target $y_i$ was not used for training $M_{\sigma(i)-1}$ (neither for the TS calculation, nor for the gradient estimation) and, second, the distribution of TS $\hat{x}^i$ conditioned by the target value is the same for a training example and a test example with the same value of feature $x^i$. 

\section{Experimental results}\label{appendix::experimental_res}

\paragraph{Comparison with baselines}
In Section~\ref{sec:experiments} we demonstrated that the strong setting of CatBoost, including ordered TS, Ordered mode and feature combinations, outperforms the baselines. Detailed experimental results of that comparison are presented in Table~\ref{tab:baselines_supplementary}.

\begin{table}[h]
\caption{Comparison with baselines: logloss / zero-one loss, relative increase is presented in the brackets.}\label{tab:baselines_supplementary}
\small
\centering
\begin{tabular}{lccc}
  \cmidrule{2-4}
 & CatBoost &  LightGBM &  XGBoost \\
   \midrule
Adult & 
\textbf{0.2695 / 0.1267}  & 
0.2760 (+2.4\%) / 0.1291 (+1.9\%) &  
0.2754 (+2.2\%) / 0.1280 (+1.0\%)  \\
Amazon & 
\textbf{0.1394 / 0.0442} & 
0.1636 (+17\%) / 0.0533 (+21\%) & 
0.1633 (+17\%) / 0.0532 (+21\%) \\
Click & 
\textbf{0.3917 / 0.1561} & 
0.3963 (+1.2\%) / 0.1580 (+1.2\%)  & 
0.3962 (+1.2\%) / 0.1581 (+1.2\%) \\
Epsilon & 
\textbf{0.2647 / 0.1086} & 
0.2703 (+1.5\%) / 0.114 (+4.1\%) & 
0.2993 (+11\%) / 0.1276 (+12\%) \\
Appetency & 
\textbf{0.0715 / 0.01768} & 
0.0718 (+0.4\%) / 0.01772 (+0.2\%) & 
0.0718 (+0.4\%)  / 0.01780 (+0.7\%) \\
Churn & 
\textbf{0.2319 / 0.0719}  & 
0.2320 (+0.1\%) / 0.0723 (+0.6\%) &  
0.2331 (+0.5\%) / 0.0730 (+1.6\%)  \\
Internet & 
\textbf{0.2089 / 0.0937} & 
0.2231 (+6.8\%) / 0.1017 (+8.6\%) &
0.2253 (+7.9\%) / 0.1012 (+8.0\%) \\
Upselling & 
\textbf{0.1662 / 0.0490}  & 
0.1668 (+0.3\%) / 0.0491 (+0.1\%)	& 
0.1663 (+0.04\%) / 0.0492 (+0.3\%) \\
Kick & 
\textbf{0.2855 / 0.0949}  & 
0.2957 (+3.5\%) / 0.0991 (+4.4\%) &
0.2946 (+3.2\%) / 0.0988 (+4.1\%)  \\
\bottomrule
\end{tabular}
\end{table}

In this section, we empirically show that our implementation of GBDT provides state-of-the-art quality and thus is an appropriate basis for building CatBoost by adding different improving options including the above-mentioned ones. For this purpose, we compare with baselines a \textit{raw setting} of CatBoost which is as close to classical GBDT~\cite{friedman2001greedy} as possible. Namely, we use CatBoost in GPU mode with the following parameters: \textit{--\,--\,boosting--type Plain\,\,
--\,--\,border--count 255\,\,
--\,--\,dev--bootstrap--type DiscreteUniform \,\,
--\,--\,gradient--iterations 1\,\,
--\,--\,random--strength 0\,\, 
--\,--\,depth 6}. 
Besides, we tune the parameters \textit{dev--sample--rate, learning--rate, l2--leaf--reg} instead of the parameters described in paragraph ``Parameter tuning'' of Appendix~\ref{sec:ExerimentalSettings} by 50 steps of the optimization algorithm. Further, for all the algorithms, all categorical features are transformed to ordered TS on the basis of a random permutation (the same for all algorithms) of training examples at the preprocessing step. The resulting TS are used as numerical features in the training process. Thus, no CatBoost options dealing with categorical features are used. As a result, the main difference of the raw setting of CatBoost compared with XGBoost and LightGBM is using oblivious trees as base predictors. 

\begin{table}[h]
\caption{Comparison with baselines: logloss / zero-one loss (relative increase for baselines).}\label{tab:Plain_vs_baselines}
\centering
\begin{tabular}{lccc}
  \cmidrule{2-4}
 & Raw setting of CatBoost &  LightGBM &  XGBoost \\
   \midrule
Adult & 
0.2800 / 0.1288  & 
-1.4\% / +0.2\% &  
-1.7\% / -0.6\%  \\
Amazon & 
0.1631 / 0.0533 & 
+0.3\% / 0\% & 
+0.1\% / -0.2\% \\
Click & 
0.3961 / 0.1581 & 
+0.1\% / -0.1\%  & 
0\% / 0\% \\
Appetency & 
0.0724 / 0.0179 & 
-0.8\% / -1.0\% & 
-0.8\%  / -0.4\% \\
Churn & 
0.2316 / 0.0718  & 
+0.2\% / +0.7\% &  
+0.6\% / +1.6\%  \\
Internet & 
0.2223 / 0.0993 & 
+0.4\% / +2.4\% &
+1.4\% / +1.9\% \\
Upselling & 
0.1679 / 0.0493  & 
-0.7\% / -0.4\%	& 
-1.0\% / -0.2\% \\
Kick & 
0.2955 / 0.0993  & 
+0.1\% / -0.4\% &
-0.3\% / -0.2\%  \\
\bottomrule
Average & 
  & 
-0.2\% / +0.2\% &
-0.2\% / +0.2\%  \\
\bottomrule
\end{tabular}
\end{table}

For the baselines, we take the same results as in  Table~\ref{tab:baselines_supplementary}. As we can see from Table~\ref{tab:Plain_vs_baselines}, in average, the difference between all the algorithms is rather small: the raw setting of CatBoost outperforms the baselines in terms of zero-one loss by 0.2\% while they are better in terms of logloss by 0.2\%. Thus, taking into account that a GBDT model with oblivious trees can significantly speed up execution at testing time~\cite{BoostedDecisionTables}, our implementation of GBDT is very reasonable choice to build CatBoost on. 

\paragraph{Ordered and Plain modes}
In Section~\ref{sec:experiments} we showed experimentally that Ordered mode of CatBoost significantly outperforms Plain mode in the strong setting of CatBoost, including ordered TS and feature combinations. In this section, we verify that this advantage is not caused by interaction with these and other specific CatBoost options. For this purpose, we compare Ordered and Plain modes in the raw setting of CatBoost described in the previous paragraph. 

In Table~\ref{tab:Ordered_vs_Plain_raw_setting}, we present relative results w.r.t. Plain mode for two modifications of Ordered mode. The first one uses one random permutation $\sigma_{boost}$ for Ordered mode generated independently from the permutation $\sigma_{cat}$ used for ordered TS. Clearly, discrepancy between the two permutations provides target leakage, which should be avoided. However, even in this setting Ordered mode considerably outperforms Plain one by 0.5\% in terms of logloss and by 0.2\% in terms of zero-one loss in average. Thus, advantage of Ordered mode remains strong in the raw setting of CatBoost.

\begin{table}[h]
\caption{Ordered vs Plain modes in raw setting: change of logloss / zero-one loss relative to Plain mode.}\label{tab:Ordered_vs_Plain_raw_setting}
\centering
\begin{tabular}{lcc}
\cmidrule{2-3}
 & Ordered, $\sigma_{boost}$ independent of $\sigma_{cat}$ &  Ordered, $\sigma_{boost}=\sigma_{cat}$ \\
   \midrule
Adult & 
-1.1\% / +0.2\%  & 
-2.1\% / -1.2\%  \\
Amazon & 
+0.9\% / +0.9\% & 
+0.8\% / -2.2\% \\
Click & 
0\% / 0\% & 
0.1\% / 0\% \\
Appetency & 
-0.2\% / 0.2\% & 
-0.5\%  / -0.3\% \\
Churn & 
+0.2\% / -0.1\%  & 
+0.3\% / +0.4\%  \\
Internet & 
-3.5\% / -3.2\% & 
-2.8\% / -3.5\% \\
Upselling & 
-0.4\% / +0.3\%  & 
-0.3\% / -0.1\% \\
Kick & 
-0.2\% / -0.1\%  & 
-0.2\% / -0.3\%  \\
\bottomrule
Average & 
-0.5\% / -0.2\%  & 
-0.6\% / -0.9\%  \\
\bottomrule
\end{tabular}
\end{table}

In the second modification, we set $\sigma_{boost}=\sigma_{cat}$, which remarkably improves both metrics: the relative difference with Plain becomes (in average) 0.6\% for logloss and 0.9\% for zero-one loss. This result empirically confirms the importance of the correspondence between permutations $\sigma_{boost}$ and $\sigma_{cat}$, which was theoretically motivated in Appendix~\ref{appendix::combination}.

\paragraph{Feature combinations}

\begin{figure}[h]
\centering
\includegraphics[width=0.7\textwidth]{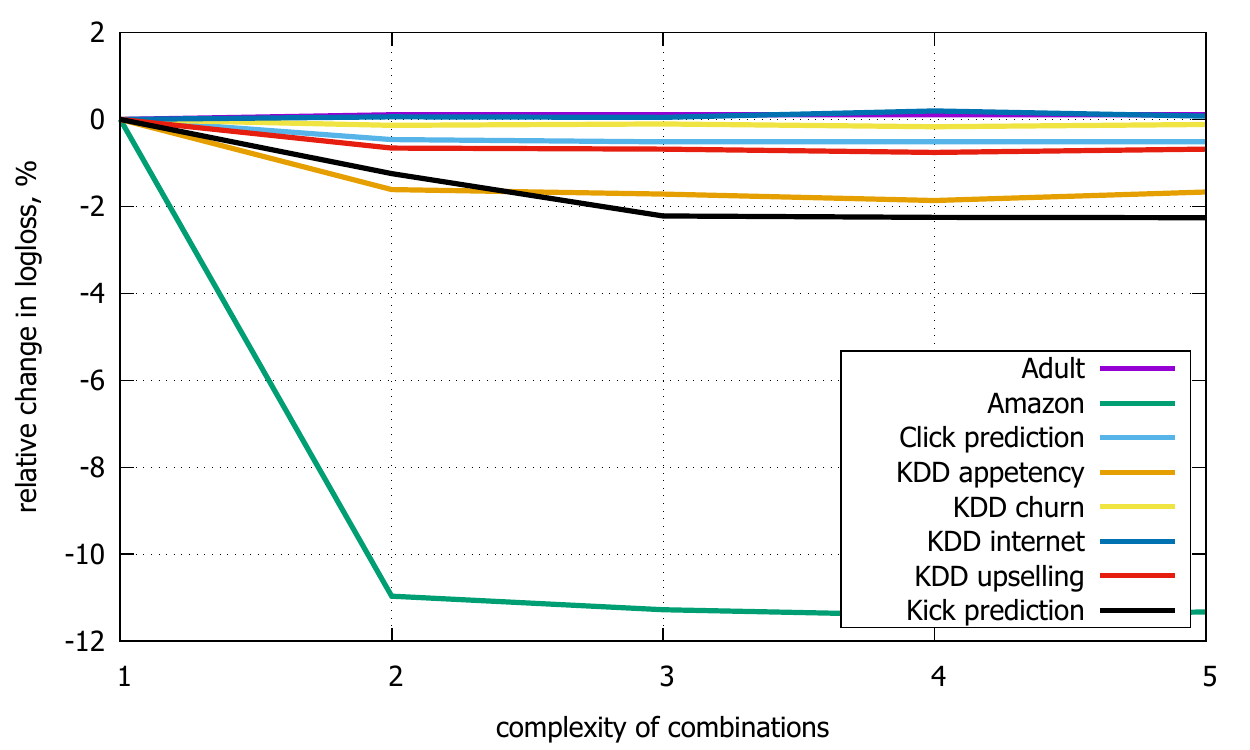}
\captionof{figure}{Relative change in logloss for a given allowed complexity compared to the absence of feature  combinations.}\label{fig:combinations}
\end{figure}

To demonstrate the effect of feature combinations, in Figure~\ref{fig:combinations} we present the relative change in logloss for different numbers $c_{max}$ of features allowed to be combined (compared to $c_{max} = 1$, where combinations are absent). In average, changing $c_{max}$ from 1 to 2 provides an outstanding improvement of $1.86\%$ (reaching $11.3\%$), changing from 1 to 3 yields $2.04\%$, and further increase of $c_{max}$ does not influences the performance significantly.
 
\begin{figure}[]
\centering
\includegraphics[width=0.7\textwidth]{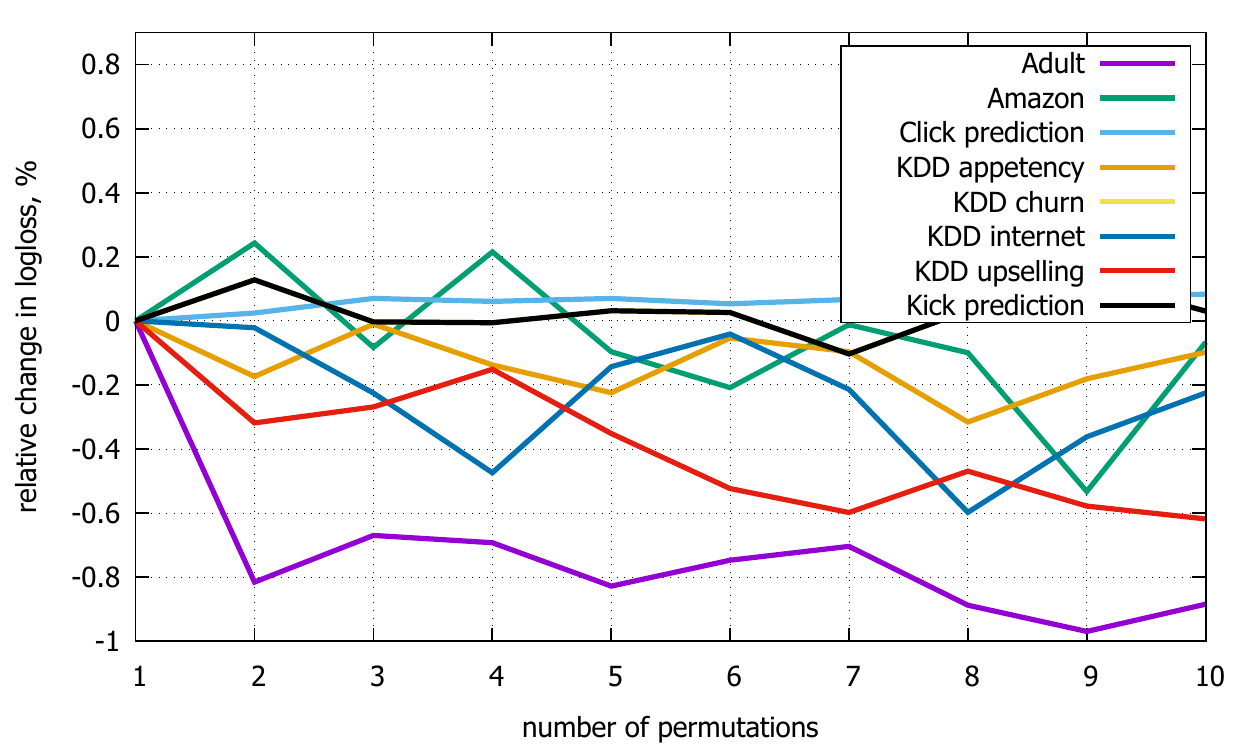}
\captionof{figure}{Relative change in logloss for a given number of permutations $s$ compared to $s=1$,}\label{fig:folds}
\end{figure}

\paragraph{Number of permutations}\label{sec:n_permut_2}
The effect of the number $s$ of permutations  on the performance of CatBoost is presented in Figure~\ref{fig:folds}.  In average, increasing $s$ slightly decreases logloss, e.g., by $0.19\%$ for $s=3$ and by $0.38\%$ for $s=9$ compared to $s=1$. 

\end{appendices}

\end{document}